\documentclass[10pt,twocolumn,letterpaper]{article}

\usepackage{multirow}
\usepackage{makecell}
\usepackage{bm}
\usepackage{xcolor}
\usepackage[pagenumbers]{cvpr} %

\usepackage{bm}
\usepackage{fvextra} %
\DefineVerbatimEnvironment{vlmprompt}{Verbatim}{
  breaklines=true, breakanywhere=true,
  fontsize=\small
}
\usepackage{adjustbox}

\definecolor{Highlight}{HTML}{39b54a}  %

\usepackage{amssymb}%
\usepackage{pifont}%
\newcommand{\cmark}{\ding{51}}%
\newcommand{\mysection}[1]{\vspace{3pt}\noindent\textbf{#1.}}

\definecolor{peach}{HTML}{F9DCBA}
\definecolor{pinkblush}{HTML}{FFE8ED}
\definecolor{lavendergray}{HTML}{E5E2F4}
\definecolor{skyblue}{HTML}{A6BEDA}
\definecolor{mintgreen}{HTML}{CBF2DC}
\definecolor{crimsonrose}{HTML}{E94B68}

\definecolor{custom_red}{RGB}{231,111,81}
\definecolor{custom_green}{RGB}{42,157,143}
\definecolor{custom_dark}{RGB}{38,70,83}
\definecolor{custom_yellow}{RGB}{233,196,106}
\definecolor{custom_orange}{RGB}{244,162,97}

\usepackage{algorithm}
\usepackage{listings}
\usepackage{etoolbox}
\makeatletter
\AfterEndEnvironment{algorithm}{\let\@algcomment\relax}
\AtEndEnvironment{algorithm}{\kern2pt\hrule\relax\vskip3pt\@algcomment}
\let\@algcomment\relax
\newcommand\algcomment[1]{\def\@algcomment{\footnotesize#1}}
\renewcommand\fs@ruled{\def\@fs@cfont{\bfseries}\let\@fs@capt\floatc@ruled
  \def\@fs@pre{\hrule height.8pt depth0pt \kern2pt}%
  \def\@fs@post{}%
  \def\@fs@mid{\kern2pt\hrule\kern2pt}%
  \let\@fs@iftopcapt\iftrue}
\makeatother

\usepackage[table]{xcolor} 
\usepackage{arydshln}
\usepackage{makecell}
\usepackage{rotating}
\usepackage{pifont}
\newcommand{\xmark}{\ding{55}}%
\newcommand{\authorsep}{\hspace{8pt}}

\definecolor{cvprblue}{rgb}{0.21,0.49,0.74}
\usepackage[pagebackref,breaklinks,colorlinks,allcolors=cvprblue]{hyperref}

\def\paperID{***} %
\def\confName{CVPR}
\def\confYear{2026}

\title{EasyV2V: A High-quality Instruction-based Video Editing Framework}

\author{
Jinjie Mai$\text{}^{1,2*}$ \authorsep Chaoyang Wang$\text{}^2$ \authorsep Guocheng Gordon Qian$\text{}^2$ \authorsep Willi Menapace$\text{}^2$ \\ \authorsep Sergey Tulyakov$\text{}^2$ \authorsep Bernard Ghanem$\text{}^1$ \authorsep Peter Wonka$\text{}^{2,1}$ \authorsep Ashkan Mirzaei$\text{}^{2*\dagger}$ \\
\normalsize $\text{}^1$ KAUST \authorsep $\text{}^2$ Snap Inc. \\
\normalsize $^*$ Core contributors \quad $^\dagger$ Project lead \\
\href{https://snap-research.github.io/easyv2v/}{\texttt{\textcolor{purple}{Project Page: \underline{https://snap-research.github.io/easyv2v/}}}}
\\
}

\begin{document}
\twocolumn[{%
\renewcommand\twocolumn[1][]{#1}%
\maketitle
\centering
\vspace{-9mm}
\includegraphics[width=\linewidth]{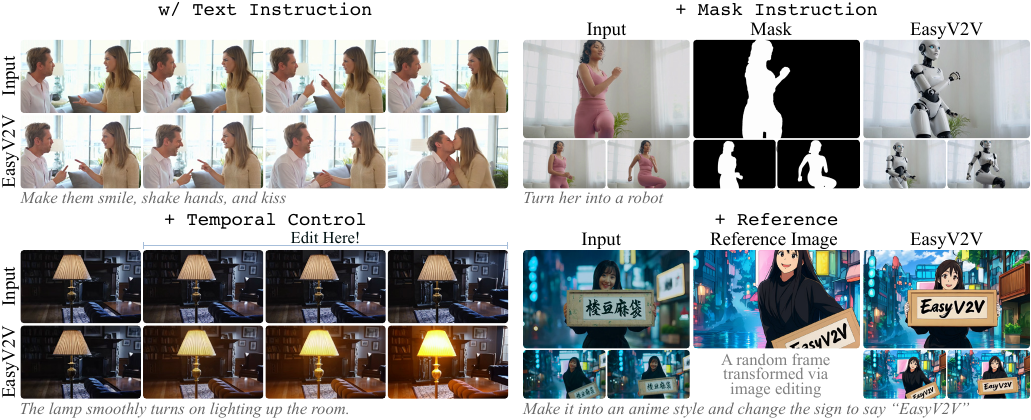}
\vspace{-7mm}
\captionof{figure}{
EasyV2V unifies data processing, architecture, and control for high-quality, instruction-based video editing with flexible inputs (text, masks, edit timing). The figure illustrates its versatility across diverse input types and editing tasks.
}
\label{fig:teaser}
\vspace{4mm}
}]

\begin{abstract}
While image editing has advanced rapidly, video editing remains less explored, facing challenges in consistency, control, and generalization.
We study the design space of data, architecture, and control, and introduce \emph{EasyV2V}, a simple and effective framework for instruction-based video editing. On the data side, we compose existing experts with fast inverses to build diverse video pairs, lift image edit pairs into videos via single-frame supervision and pseudo pairs with shared affine motion, mine dense-captioned clips for video pairs, and add transition supervision to teach how edits unfold.
On the model side, we observe that pretrained text-to-video models possess editing capability, motivating a simplified design. Simple sequence concatenation for conditioning with light LoRA fine-tuning suffices to train a strong model.
For control, we unify spatiotemporal control via a single mask mechanism and support optional reference images.
Overall, EasyV2V works with flexible inputs, e.g., video+text, video+mask+text, video+mask+reference+text, and achieves state-of-the-art video editing results, surpassing concurrent and commercial systems. 
\end{abstract}
    
\vspace{-10mm}
\section{Introduction}
\label{sec:intro}

What makes a good instruction-based video editor?
We argue that three components govern performance: \emph{data}, \emph{architecture}, and \emph{control}.
This paper analyzes the design space of these components and distills a recipe that works great in practice.
The result is a lightweight model that reaches state-of-the-art quality while accepting flexible inputs.

\begin{figure}
\centering
\includegraphics[width=1.0\linewidth]{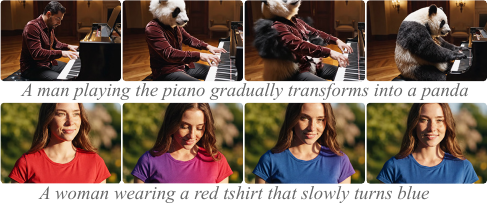}
\vspace{-7mm}
\caption{A pretrained text-to-video model can mimic common editing effects without finetuning. This suggests that much of the ``how'' of video editing already lives inside modern backbones. 
}
\label{fig:motivation}
\vspace{-5mm}
\end{figure}

Training-free video editors adapt pretrained generators but are fragile and slow~\cite{couairon2024videdit,ku2024anyv2v}.
Training-based approaches improve stability, yet many target narrow tasks such as ControlNet-style conditioning~\cite{VideoX-Fun2025,vace}, video inpainting~\cite{zi2025minimax}, or reenactment~\cite{wananimate2025}.
General instruction-based video editors handle a wider range of edits~\cite{cheng2023consistentvideotovideotransferusing,zi2025senorita,wu2025insvie,yu2025veggie}, yet still lag behind image-based counterparts in visual fidelity and control. We set out to narrow this gap.
Figure~\ref{fig:motivation} motivates our design philosophy: modern video models already know how to transform videos.
To unlock this emerging capability with minimal adaptation, we conduct a comprehensive investigation into data curation, architectural design, and instruction-control.

\textbf{Data.}
We see three data strategies.
\emph{(A) One generalist model} renders all edits and is used to self-train an editor~\cite{cheng2023consistentvideotovideotransferusing,wu2025insvie,yu2025veggie}. This essentially requires a single teacher model that already solves the problem in high quality.
\emph{(B) Design and train new experts} for specific edit types, then synthesize pairs at scale~\cite{zi2025senorita}.
This yields higher per-task fidelity and better coverage of hard skills, but training and maintaining many specialists is expensive and slows iteration and adaptation to future base models.
We propose a new strategy: \emph{(C) Select existing experts} and compose them.
We focus on experts with a fast inverse (e.g., edge$\leftrightarrow$video, depth$\leftrightarrow$video) and compose more complex experts from them. 
This makes supervision easy to obtain, and experts are readily available as off-the-shelf models.
It keeps costs low and diversity high by standing on strong off-the-shelf modules; the drawback is heterogeneous artifacts across experts, which we mitigate through filtering and by favoring experts with reliable inverses.
Concretely, we combine off-the-shelf video/image experts for stylization, local edits, insertion/removal, and human animation; we also convert image edit pairs into supervision via two routes: single-frame training and pseudo video-to-video (V2V) pairs created by applying the same smooth camera transforms to source and edited images.
Lifting image-to-image (I2I) data increases scale and instruction variety; its limitation is weak motion supervision, and shared camera trajectories reintroduce temporal structure without changing semantics.
We further leverage \emph{video continuation}: we derive V2V pairs from densely captioned text-to-video (T2V) datasets by sampling input clips outside each captioned interval and target clips within it, while converting the caption into an instruction using an LLM.
Continuation data teaches action and transition edits that are scarce in typical V2V corpora, at the cost of more careful caption-to-instruction normalization.
We collect and create an extensive set of V2V pairs, the most comprehensive among published works, and conduct a detailed study of how editing capability emerges from each training source.

\textbf{Architecture.}
Using a pretrained video backbone, we study two strategies to inject the source video: \emph{channel concatenation} and \emph{sequence concatenation}.
Channel concatenation is faster in practice because it uses fewer tokens, but we show that sequence concatenation consistently yields higher edit quality.
The trade-off is efficiency versus separation: channel concatenation keeps context short but entangles source and target signals; sequence concatenation costs more tokens but preserves clean roles for each stream, improving instruction following and local detail.
Our final design adds small, zero-init patch-embedding routes for the source video and an edit mask used for spatiotemporal control of the edit extent, reuses the frozen video VAE for all modalities, and fine-tunes only with LoRA~\cite{hu2022lora}.
Full finetuning can help when massive, heterogeneous data are available, but it risks catastrophic forgetting and is costly to scale.
With $<10$M paired videos, LoRA transfers faster, reduces overfitting, and preserves pretrained knowledge while supporting later backbone swaps; its main downside is a slight headroom loss when unlimited data and compute are present.
We inject the mask by token addition, and concatenate tokens from the source and optional reference image along the sequence.
This preserves pretraining benefits, keeps token budgets tight (by not introducing new tokens for the mask), and makes the model easily portable to future backbones.
Addition for masks is simple and fast; while it carries less capacity than a dedicated mask-token stream, we find it sufficient for precise region and schedule control without context bloat.
We use an optional reference frame at training and test time to leverage strong image editors when available.
References boost specificity and style adherence when present; randomly dropping the reference during training keeps the model robust when they are absent or noisy.

\textbf{Flexible control.}
Prior work explores control by skeletons, segmentation, depth, and masks.
A key signal is still missing: \emph{when} the edit happens.
Users often want an edit to appear gradually (e.g., “set the house on fire starting at 1.5s, then let the flames grow gradually”).
We unify spatial and temporal control with a single mask video.
Pixels mark \emph{where} to edit; frames mark \emph{when} to edit, and how the effect evolves.
Alternatives include keyframe prompts or token schedules, which are flexible but harder to author and to align with motion.
A single mask video is direct, differentiable, and composes well with text and optional references.
The cost is requiring a mask sequence, which we keep lightweight and editable.

Combining these design choices results in a unified editor, \textbf{EasyV2V}, which supports flexible input combinations, including {\texttt{video + text}}, {\texttt{video + mask + text}}, and {\texttt{video + mask + reference + text}} (see Fig.~\ref{fig:teaser}).
Despite its simplicity, our recipe consistently improves edit quality, motion, and instruction following over recently published methods.
In summary, our contributions are:
\begin{itemize}
\item A clarified design space for instruction-based video editing and a consistent strategy across data, architecture, and control that achieves state-of-the-art results.
\item A reusable data engine built from composable experts with trivial inverses, lifted high-quality image edits, and video continuation, with per-expert/per-source ablations.
\item A lightweight architecture that minimally modifies a pretrained video backbone: zero-init patch-embeddings for source and mask, frozen VAE reuse, and LoRA finetuning, plus optional reference frames.
\item \emph{Temporal control} as a first-class signal, unified with spatial control via a single mask video that schedules when edits start and how they evolve.
\end{itemize}

\section{Related work}
\label{sec:related_work}

\noindent\textbf{Instruction-based visual editing datasets.}
With the success of image and video generative models trained on large-scale text-to-image and text-to-video datasets, recent work has focused on developing datasets for instruction-based image and video editing. Early approaches attempted to build large-scale paired I2I datasets, though with low success rates, requiring extensive automatic~\cite{brooks2023instructpix2pix,hui2024hqedit} or manual~\cite{magicbrush2023} filtering. Later efforts improved success rates by leveraging task-specific models~\cite{zhao2024ultraedit,wei2025omniedit,yu2025anyedit,ge2024seed,yang2024editworld}, leading to highly capable image editing systems~\cite{openai2024gpt4ocard,labs2025flux1kontextflowmatching}. These models subsequently enabled the generation of higher-quality paired I2I datasets~\cite{gpt_image_edit}.
Creating paired V2V datasets is inherently more challenging, as it requires editing multiple frames while maintaining temporal coherence and faithfully applying the intended modification. Early work~\cite{cheng2023consistentvideotovideotransferusing} used LLMs and Prompt-to-Prompt~\cite{hertz2022prompt} to synthesize video pairs, but results were limited by artifacts from the underlying editing method. 
Another dataset~\cite{hu2025vivid10mdatasetbaselineversatile} provided video–object-mask pairs with corresponding captions but omitted edited videos. 
The current trend is to generate synthetic datasets using one previous general video editing model~\cite{cheng2023consistentvideotovideotransferusing,wu2025insvie,yu2025veggie} or a collection of more specialized video editing models~\cite{zi2025senorita}. These efforts still lag behind image editing datasets in quality and diversity. We discuss strategies to mitigate this gap and provide extensive analysis comparing the effects of different approaches.

\noindent\textbf{Instruction-based visual editing.}
Building on the success of diffusion models for image~\cite{Rombach_2022_CVPR,flux2024,gao2025seedream30technicalreport,imagenteamgoogle2024imagen3,openai2024gpt4ocard,seedream2025seedream40nextgenerationmultimodal} and video~\cite{videoworldsimulators2024,kong2024hunyuanvideo,genmo2024mochi,yang2024cogvideox,wan2025wanopenadvancedlargescale} generation, early visual editing methods used pretrained diffusion models in a training-free manner for image editing~\cite{hertz2022prompt,cao2023masactrl,ju2023direct,wang2024vid2vidzero,qi2023fatezero,liu2024videop2p,yoon2025raccoon,kulikov2024flowedit}, typically manipulating attention maps and latent spaces, or by noise inversion. These approaches generally produce low-quality outputs, have slow inference, and exhibit low success rates. Subsequent work showed that even relatively low-quality paired datasets can outperform training-free methods~\cite{brooks2023instructpix2pix,sheynin2024emuedit}. Such models concatenate input image latents with noisy latents along the channel dimension and fine-tune a pretrained image generator on synthetic paired I2I data produced by training-free approaches.
Another line of research performs concatenation along the sequence dimension, improving quality at the cost of efficiency~\cite{ju2025fullditmultitaskvideogenerative,xiao2025omnigen,hidreami1technicalreport,zhang2025icedit}. Similar directions adopt LLM-style architectures for unified or token-based editing~\cite{deng2025bagel,zhou2024transfusionpredicttokendiffuse}.
Instruction-based video editing has been explored far less. Early attempts include training-free video editing pipelines~\cite{couairon2024videdit,ku2024anyv2v}. Some works adapt image generative models~\cite{singer2024eve, InstructVid2Vid}, while others train on low-quality synthetic video pairs and are limited by training data quality~\cite{hertz2022prompt,cheng2023consistentvideotovideotransferusing,zhang2024effived,yu2025veggie}. Additional methods include propagating edits from the first frame to subsequent frames~\cite{liu2025genprop} or designing architectures with task-specific conditioning for multiple editing tasks~\cite{ye2025unic}. 
Another line of work explores task-specific editing, such as ControlNet-style~\cite{zhang2023adding} video generation~\cite{vace} or pose/face conditioned human video synthesis~\cite{wananimate2025}.
Two concurrent frameworks recently appeared on arXiv, focusing on patch-wise concatenation~\cite{lucyedit2024} and LLM-style architectures~\cite{editverse2025}. In contrast, our method systematically studies different data sources for training V2V models, identifies those most critical for performance, and enables more controllable video generation through optional reference images and spatiotemporal edit control.

\section{Method}
\label{sec:method}

\begin{figure*}
    \centering
    \includegraphics[width=1.0\linewidth]{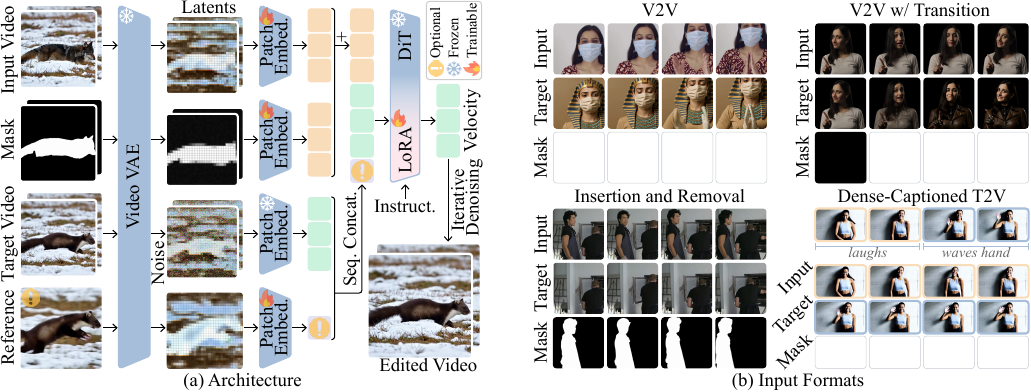}
    \vspace{-7mm}
    \caption{(a) Overview of our video editing architecture. The frozen video VAE encodes control signals. Mask tokens are added to the input video tokens and concatenated with the noisy latent tokens and optional reference image. The DiT is trained using LoRA.
    (b) Input formats used for different dataset types during training. When object masks are available, the spatiotemporal mask input equals the input mask. Otherwise, the mask indicates transitions from the unedited to the edited video. }
    \label{fig:method}
    \vspace{-3mm}
\end{figure*}

\mysection{Overview}
We build on a modern video generative backbone~\cite{wan2025wanopenadvancedlargescale} and introduce lightweight conditioning modules for video editing.
A pivotal design choice is how to inject the source video condition. While one simple approach is to concatenate the source video latent and the noisy latent along the channel dimension~\cite{lucyedit2024}, requiring only a modification of the initial patch embedding layer, we found this method struggles to learn edits efficiently. We therefore adopt a sequence-wise concatenation strategy, appending the source video tokens to the noisy latent token sequence. This approach provides a more robust and effective conditioning mechanism, as validated in our ablations (Tab.~\ref{tab:arch_ablation}).

Concretely, we add separate patch-embedding layers for each control signal (source video and mask) and support an optional reference image at training and inference. 
After the encoding of inputs and patch-embedding, condition signals are injected either by element-wise addition (for the edit mask) or by sequence concatenation (for source video and optional reference image).
To preserve the generative priors of the T2V backbone and ensure stable training, we adopt a parameter-efficient finetuning strategy. We freeze the original backbone weights and optimize only the newly introduced patch-embedding layers and low-rank adaptation (LoRA) weights added to the DiT's attention layers. We found this approach essential, as full-model finetuning easily led to training instability and source video inconsistency.

\mysection{Input video conditioning}
As shown in Fig.~\ref{fig:method}, we adopt a simple yet effective strategy to condition on the input video. 
Given a source video and a target video, each of shape $V \in \mathbb{R}^{NCHW}$, we encode them with a video VAE into latents $Z_{\text{src}}, Z_{\text{tgt}} \in \mathbb{R}^{nchw}$. 
We use \emph{separate} patch-embedding layers for $Z_{\text{src}}$ and $Z_{\text{tgt}}$.
After applying edit-mask conditioning (below), their tokens are concatenated along the sequence dimension. 
We place $Z_{\text{src}}$ first, followed by $Z_{\text{tgt}}$, so the model can learn in-context video editing behavior that resembles video continuation, which is supported by our dense-caption video data.

\mysection{Edit mask conditioning}
The mask video $M \in \mathbb{R}^{NCHW}$ is a binary indicator of where and when to edit the source video. 
It can specify (1) pixel-wise regions for inpainting or removal and/or (2) frame-wise intervals to control when an edit occurs and for how long, enabling both spatial and temporal controllability. 
We process the encoded mask latents $Z_{\text{msk}}$ with a dedicated patch-embedding layer and inject them by addition into the source video tokens. 
This addition-based injection is a deliberate choice for computational efficiency. 
Because the mask is a low-frequency signal, its information can be effectively fused without appending it to the DiT's input sequence. 
At inference time, if no mask video is provided, we default to blank masks, treating the task as a canonical, instruction-only video edit. 

\mysection{Reference image conditioning}
We optionally provide a reference image during both training and inference. 
The reference can be sampled from the target video during training. 
During inference, it can be produced by an external image-editing model by applying it to a frame from the source video or provided by the user. 
Because references can be imperfect (e.g., spurious zooms from Qwen-Image-Edit~\cite{qwenimage2025}), we apply random crops/rotations and randomly drop the reference during training so the model is robust to its absence or noise at test time.
When present, the reference is encoded with the video VAE and $\texttt{Patch\_embedding}_{\text{ref}}$, and its tokens are concatenated at the end of the sequence. 
This preserves a fixed token distance between $Z_{\text{src}}$ and $Z_{\text{tgt}}$ while positioning $Z_{\text{ref}}$ closer to $Z_{\text{tgt}}$ for stronger guidance, leveraging off-the-shelf image-editing priors without sacrificing inference flexibility.

\section{Data pipeline}
\label{sec:data_pipeline}

\begin{figure*}
    \centering
    \includegraphics[width=0.95\linewidth]{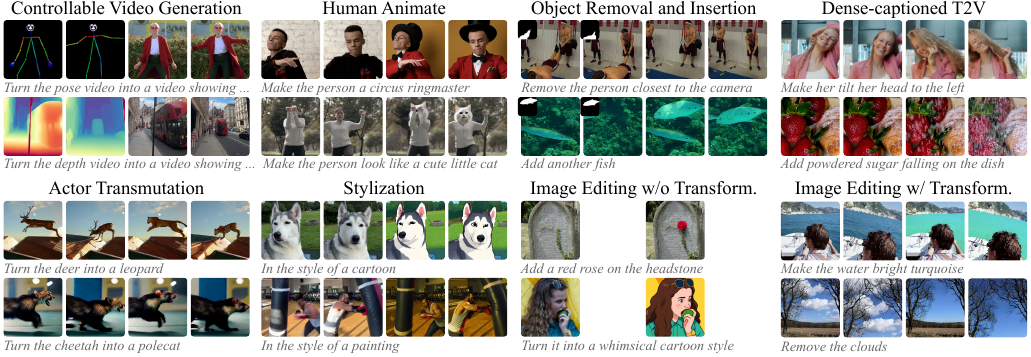}
    \vspace{-3mm}
    \caption{Overview of our datasets. In each example, the left images are inputs and the right images are outputs.}
    \label{fig:dataset}
    \vspace{-3mm}
\end{figure*}

\begin{table}[t]
\centering
\caption{Summary of datasets used for training. We collect $\sim4.3$M open-sourced and licensed data and curate $\sim3.4$M data, resulting in a total of $\sim8$M high-quality data for training. }
\vspace{-3mm}
\label{tab:dataset_summary}
\resizebox{\columnwidth}{!}{%
\begin{tabular}{@{}c lcccl@{}}
\toprule
& \textbf{Dataset} & \textbf{Type} & \textbf{Mask} & \textbf{Reference} & \textbf{\# Pairs} \\ 
\midrule
\multirow{2}{*}{\rotatebox{90}{\textbf{}}} 
& GPT-Edit-1.5M~\cite{gpt_image_edit} & I2I & \xmark & \xmark & $\sim1.5$M \\
& Ditto-1M~\cite{bai2025ditto} & V2V & \xmark & \xmark &  $\sim1$M \\
& Señorita-2M~\cite{zi2025senorita} & V2V & \xmark & \xmark & $\sim1.8$M \\
\hdashline
\multirow{7}{*}{\rotatebox{90}{\textbf{Ours}}}
& Image Editing & I2I & \xmark & \xmark &  $\sim2$M \\
& Human Animate & V2V & \xmark & \cmark &  $\sim60$K \\
& Object Removal / Insertion & V2V & \cmark & \xmark &  $\sim110$K \\
& Actor Transmutation & V2V & \xmark & \cmark &  $\sim4$K \\
& Video Stylization & V2V & \xmark & \cmark &  $\sim90$K \\
& Controllable Video Generation & V2V & \xmark & \xmark &  $\sim1.1$M \\
& Human Action & T2V & \xmark & \cmark &  $\sim150$K \\
\bottomrule
\end{tabular}%
}
\vspace{-3mm}
\end{table}

\subsection{V2V data}

We propose a data generation framework that utilizes various expert pipelines for generating paired V2V datasets. Our approach yields significantly stronger results than only using a single general editing pipeline. Figure~\ref{fig:dataset} shows samples of each of our new datasets. Next, we detail the experts. 

\noindent\textbf{Human animation.}
Previous V2V datasets~\cite{zi2025senorita} lack expert annotations ensuring consistent human poses and facial expressions between input and edited videos. To overcome this, we use Wan Animate~\cite{wananimate2025} to generate paired V2V data where the output video is conditioned on (1) the subject’s pose in the input video for pose consistency, (2) cropped facial regions for expression consistency, and (3) a reference image produced by an image editing model that modifies the first frame to alter the subject or scene. A large language model (LLM) generates diverse edit instructions involving actor swaps, clothing and style changes, and the addition or removal of accessories.

\noindent\textbf{Object removal and insertion.}
We construct paired videos by removing selected objects from original clips. First, we apply an open-set object detection model~\cite{zhang2023recognize} to the first frame of each video to detect candidate objects and generate corresponding bounding boxes and tags. Next, an LLM refines the tags by removing adjectives (e.g., color, material) and scenery-related labels (e.g., road, sky). For each video, we randomly sample a fixed number of objects, with sampling probability proportional to the bounding box area to prioritize salient objects. For each selected object, we use a video segmentation model~\cite{ravi2024sam2} to obtain per-frame segmentation masks. A video inpainting model~\cite{zi2025minimax} then generates a new version of the video with the target object removed. Human annotators review the results and discard videos with noticeable inpainting artifacts. Finally, we provide the input and output videos, along with the object category and input video captions, to a vision-language model (VLM)~\cite{bai2025qwen25vltechnicalreport} and prompt it to generate a detailed instructional caption describing the edit. By replacing the source and target videos and corresponding VLM-based changes to the instruction, this dataset is used for object insertion.

\noindent\textbf{Actor transmutation.}
While the human animate dataset already includes high-quality examples of replacing one humanoid with another, a general editing model should also support actor transmutation for other categories such as animals. To achieve this, we leverage a zero-shot image editing model~\cite{kulikov2024flowedit} and build a zero-shot V2V pipeline on top of a pretrained video generative model~\cite{wan2025wanopenadvancedlargescale}. We use this zero-shot V2V model to extend editing types beyond humanoid-to-humanoid, including (1) quadruped-to-quadruped, (2) bipedal-to-bipedal, and (3) avian-to-avian transformations. We observe that the zero-shot editing model~\cite{kulikov2024flowedit} performs noticeably better when the input video is also generated by the same video generative model. Consequently, for each video pair, we sample two actors from the same category, along with an action and a scene, and generate a video of ``\textit{\textless actor1\textgreater\; performing \textless action\textgreater\; in \textless scene\textgreater\;}". We then apply our zero-shot V2V pipeline to produce the edited version described as ``\textit{\textless actor2\textgreater\; performing \textless action\textgreater\; in \textless scene\textgreater\;}". The lists of actors, actions, and scenes for each category are generated by an LLM.

\noindent\textbf{Video stylization.}
A general-purpose V2V model should be capable of applying global style changes to input videos. To create suitable datasets for this task, we extract edge maps from input videos to preserve structure and motion while removing appearance and style information. We then use an image editing model~\cite{labs2025flux1kontextflowmatching} to apply an  LLM generated style transfer instruction to the first frame of the input video. This edited reference image, along with the edge video, are used as control signals to generate the stylized output video~\cite{VideoX-Fun2025}. The style transfer instructions include art movements and styles, lighting conditions, aesthetics, photographic and cinematic styles, weather conditions, traditional arts and patterns, and color palettes and tones.

\noindent\textbf{Controllable video generation.}
Video generation using control signals such as depth and edge maps has been widely studied~\cite{VideoX-Fun2025,vace}. The datasets for training such models consist of video pairs that are simple and inexpensive to obtain, making them suitable as supporting data for training V2V models. These datasets enhance controllable generation capabilities and potentially promote emergent skills through data diversification. We include controllable video generation pairs in our training data, using control signals such as depth, HED edge, Canny edge, optical flow, human poses, noisy videos, and grayscale videos. 

\noindent\textbf{V2V transition data.}
To supervise \emph{how} an edit unfolds over time and enable temporal mask control, we synthesize transition effects on top of paired videos $V^{\text{src}}$ and $V^{\text{tgt}}$. 
Given an edit onset $t_i$, we form a training target $V' = [\,V^{\text{src}}_{t_0:t_i},\, V^{\text{tgt}}_{t_i:t_N}\,]$ and derive a frame-wise mask that activates the edit after $t_i$. 
We then apply transition operators like linear blending to ensure a natural transition effect centered at $t_i$.

\subsection{I2I data}
High-quality instructional V2V datasets are limited, while image editing benefits from mature models and large-scale data. 
 To bridge this gap, we train our instructional V2V editor with I2I edit pairs collected from open-source datasets~\cite{gpt_image_edit,zhao2024ultraedit,wei2025omniedit,hui2024hqedit} and additional pairs synthesized from VLM-curated image–caption corpora. From each caption, an LLM produces diverse instruction-style edit prompts; instruction-following image editors~\cite{qwenimage2025,labs2025flux1kontextflowmatching} generate the edited images. A VLM-based filter retains only successful, high-quality edits, yielding instruction-aligned I2I supervision without manual annotation.

\noindent\textbf{I2I $\rightarrow$ V2V via affine transformations.}
Treating an I2I pair as a single-frame video is insufficient because it lacks temporal and motion cues. To address this, we convert each I2I pair into a \emph{pseudo} video pair using a shared 2D affine camera trajectory. Specifically, we sample smooth sequences of small rotations, zooms, and translations by interpolating between a neutral and a randomly drawn target pose, with adjustments to prevent out-of-frame regions. The same trajectory is applied to both the source and target images, producing temporally consistent videos that differ only by the intended edit. This preserves the I2I supervision signal while introducing realistic temporal structure.

\subsection{Dense-captioned text-to-video data}
Existing V2V datasets cover object manipulation, stylization, and animation but lack edits like changing \emph{human action}. As a result, instruction-guided V2V models often fail to modify actions effectively. Dense-captioned text-to-video datasets~\cite{mint}, however, provide diverse action descriptions at scale, enabling action-centric supervision.
To curate T2V data for video editing, given a video $V$ and a caption $c$ localized to a temporal window $[t_i, t_j]$ (e.g., \emph{“he sits down”}), we slice $V$ to form a \emph{source} clip $V_{t_{i-N}:t_i}$ (frames preceding the caption) and a \emph{target} clip $V_{t_i:t_{i+N}}$ (frames containing the caption), where $N$ is the frame count used for training. We discard windows that are too long ($t_j - t_i \gg N$), too short, or contain scene cuts. The caption $c$ is then converted into an imperative instruction $c'$ using an LLM~\cite{qwen3} (e.g., \emph{“make him sit down”}). This yields canonical triples for training V2V editors: \texttt{(source video, target video, instruction)}. Our approach scales efficiently to large captioned video corpora while emphasizing action edits.
Qualitative results (Figure~\ref{fig:teaser}) reveal that curated action edits enable EasyV2V to perform accurate and complex human action edits.

\section{Implementation details}

We adopt the pretrained video generation model, Wan-2.2-TI2V-5B~\cite{wan2025wanopenadvancedlargescale}, as our base model for LoRA adaption and Wan-2.2-VAE with a spatiotemporal compression ratio of $4\times16\times16$.
We perform training and report our results based on a spatiotemporal resolution of $81\times832\times480$ following Wan-2.1-FunControl.
We also provide high-quality results fine-tuned on $81\times1280\times704$ in the supplement.
For LoRA training, we set the rank to $256$ and adopt a constant learning rate of $1e^{-4}$ with AdamW~\cite{adamw} optimizer.
We conduct our training on 32 NVIDIA H100s for our EasyV2V.
All newly introduced parameters are zero-initialized.
We perform random reference image dropout and video transition augmentation each with $50\%$ probability.
A complete list of all training data used is provided in Tab.~\ref{tab:dataset_summary}.

\section{Experiments}
\label{sec:experiments}

\subsection{Benchmarks and metrics}

We evaluate our method on the latest EditVerseBench~\cite{editverse2025} which has 20 edit types.
We evaluate on categories covered by our training data—excluding unsupported tasks such as camera-pose changes—resulting in 160 videos across 16 edit types.
Following the benchmark's protocol, we conduct a comprehensive quantitative evaluation using these metrics:  
\textbf{Frame and video text alignment.} We measure alignment at both frame and video levels using image-text~\cite{clip2021} and video-text~\cite{viclip} encoders by computing cosine similarity between (i) each frame and the target prompt and (ii) a joint embedding of uniformly sampled frames and the prompt.  
\textbf{Preference score.} We report a preference-tuned image-text PickScore~\cite{pickscore} that correlates with human aesthetic judgments, applied between frames and the target prompt.  
\textbf{VLM quality assessment.} We use OpenAI GPT-4o~\cite{openai2024gpt4ocard} to evaluate three sampled frame pairs on prompt following, edit quality, and background consistency (0-3 each), providing justifications and a total score of $9$.
In our analysis, we observe that the VLM score aligns most closely with human qualitative assessment so we designate the VLM score as our primary metric for comparison.

\subsection{Baseline comparison}

We perform a comprehensive quantitative comparison against a diverse set of baseline methods in Table~\ref{tab:main_results}. 
With our primary VLM score of $7.73/9$, without any guidance, we outperform not only the previously best published method but also concurrent work and a commercial solution.
When a reference image is given, we can achieve better visual-text alignment performance.

\begin{figure*}
    \centering
    \includegraphics[width=0.9\linewidth]{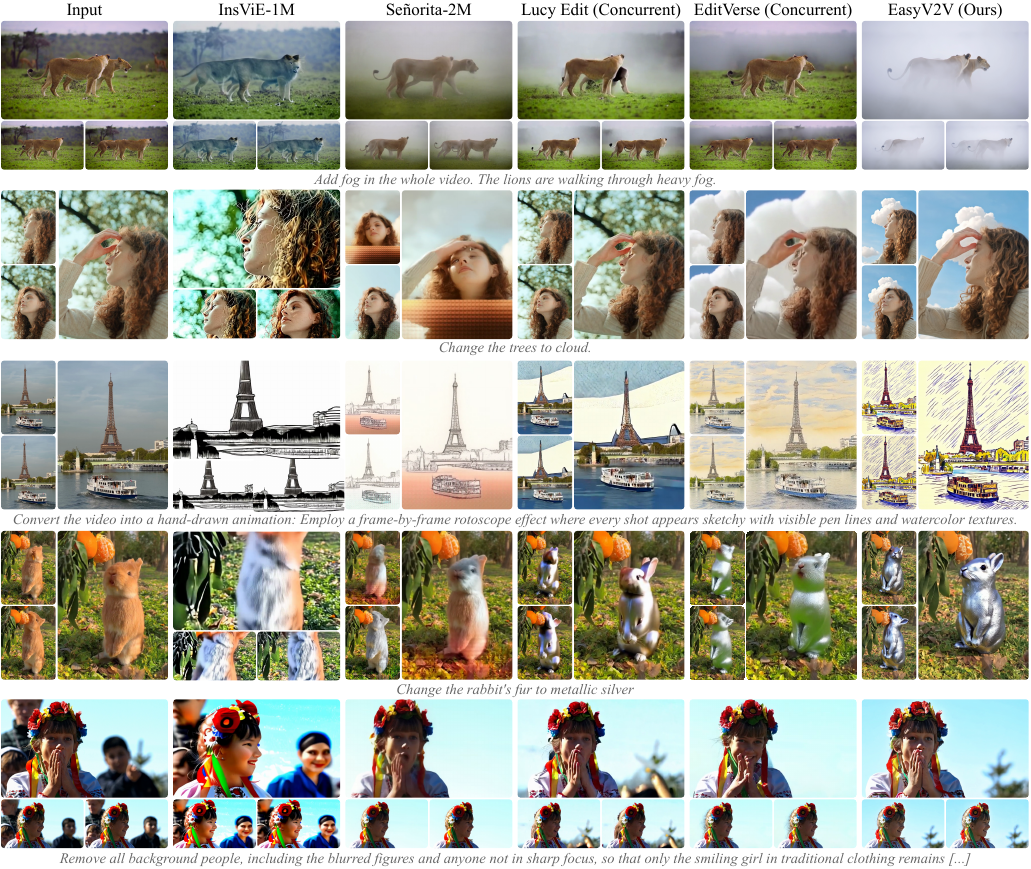}
    \vspace{-3mm}
    \caption{Qualitative comparison of our video editing results with the baselines.}
    \label{fig:results_comparison}
    \vspace{-3mm}
\end{figure*}

We provide a qualitative comparison with the baselines in Fig.~\ref{fig:results_comparison}. Since EditVerse~\cite{editverse2025} has not released its code, we take their gallery videos from their webpage. 
InsViE-1M~\cite{wu2025insvie} supports only short horizontal videos and often fails to edit properly, producing severe visual artifacts.  
Señorita-2M~\cite{zi2025senorita} depends on the image editor for the first frame and, even with successfully edited first frames, shows motion mismatch and artifacts after the first frame. 
Lucy Edit Dev~\cite{lucyedit2024} supports limited categories of edit types and frequently exhibits motion mismatches.
Compared to these methods, our model achieves higher-quality outputs and better instruction following. For example, in the first row, EditVerse fails to produce a ``heavy fog'' across the ``whole video.'' In the second, the tree branch between her fingers remains. In the third, the output lacks ``visible pen lines.'' In the last, EditVerse also removes the background trees. Please refer to our supplement for visualizations.

\begin{table}[t]
\centering
\caption{Benchmark comparison across methods on EditVerse Benchmark. Higher is better for all metrics (↑).
}
\vspace{-3mm}
\label{tab:main_results}
\resizebox{\columnwidth}{!}{%
\begin{tabular}{lcccc}
\toprule
\multirow{2}{*}{\textbf{Method}} &
\multirow{2}{*}{\begin{tabular}[c]{@{}c@{}}\textbf{VLM evaluation}\\ Editing Quality $\uparrow$\end{tabular}} &
\multirow{2}{*}{\begin{tabular}[c]{@{}c@{}}\textbf{Video Quality}\\ Pick Score $\uparrow$\end{tabular}} &
\multicolumn{2}{c}{\textbf{Text Alignment} $\uparrow$} \\
& & & Frame & Video\\
\midrule
\multicolumn{5}{l}{\textit{Attention Manipulation (Training-free)}} \\
TokenFlow~\cite{tokenflow}      & 5.02 & 19.59 & 25.10 & 22.49 \\
STDF~\cite{STDF}           & 4.20 & 19.32 & 24.74 & 22.09 \\
\hdashline
\multicolumn{5}{l}{\textit{Instruction-Guided (w/ End-to-End Training)}} \\
Señorita-2M~\cite{zi2025senorita} (w/ Ref., Qwen-Image-Edit)   & 6.45 & 20.26 & 26.51 & 23.24 \\
InsViE-1M~\cite{wu2025insvie} (w/ Ref)   & 4.36 & 19.25 & 25.06 & 21.28 \\
InsV2V~\cite{cheng2023consistentvideotovideotransferusing}         & 4.95 & 19.33 & 24.98 & 22.74 \\
\hdashline
Ours (w/o Ref.)              & \cellcolor{mintgreen!100}7.73 & 20.36 & \cellcolor{peach!50}27.59 & \cellcolor{peach!50}24.46 \\
Ours (w/ Ref., Qwen-Image-Edit)   & 7.36 & \cellcolor{peach!50}20.54 & 27.50 & 24.31 \\
Ours (w/ Ref., Flux-Kontext)              &  \cellcolor{peach!50}7.53 & \cellcolor{mintgreen!100}20.61 & \cellcolor{mintgreen!100}28.10 & \cellcolor{mintgreen!100}25.13 \\

\hdashline
\multicolumn{5}{l}{\textit{Closed-Source Commercial Models}} \\
Runway Aleph~\cite{runway2025aleph}   & \textcolor{gray!60}{7.48} & \textcolor{gray!60}{20.56} & \textcolor{gray!60}{27.96} & \textcolor{gray!60}{24.68} \\

\hdashline
\multicolumn{5}{l}{\textit{Concurrent unpublished work}} \\
Lucy Edit~\cite{lucyedit2024}      & \textcolor{gray!60}{5.96} & \textcolor{gray!60}{19.64} & \textcolor{gray!60}{26.03} & \textcolor{gray!60}{23.37} \\
EditVerse~\cite{editverse2025} (code unavailable) & \textcolor{gray!60}{7.64} & \textcolor{gray!60}{20.33} & \textcolor{gray!60}{27.70} & \textcolor{gray!60}{25.37}\\
\bottomrule
\end{tabular}%
}
\vspace{-2mm}
\end{table}

\subsection{Ablation study}

\mysection{Architecture}
We ablate the architecture choice for input video conditioning and compare full model finetuning with LoRA.  
We evaluate the VLM score for each protocol trained with $20$K and $40$K steps in Tab.~\ref{tab:arch_ablation}.  
We show that full model training tends to overfit, while LoRA tuning quickly transforms a T2V model into a V2V model.  
And \textit{``EmbedAdd.''} refers to adopting our patch embedding addition strategy for both mask video latent and control video latent. 
Technically, this is similar to channel-wise concatenation of source video latent and noisy target latent except for the bias term.  
However, our results show that this remains suboptimal compared to sequence-wise concatenation of control latents.  
We also show that LoRA tuning via patch embedding addition performs worse than sequence concatenation.
\begin{table}[t]
\centering
\caption{Ablation on architecture choice. \textit{``Full''} denotes full model training, and \textit{``EmbedAdd.''} refers to summing control and target latents after patch embedding. }
\vspace{-3mm}
\label{tab:arch_ablation}
\resizebox{0.9\columnwidth}{!}{%
\begin{tabular}{lcc}
\toprule
\textbf{Method} & \textbf{VLM score @ 20K} $\uparrow$ & \textbf{VLM score @ 40K} $\uparrow$ \\
\midrule
Full w/ EmbedAdd.                 &    4.67    &   4.57     \\
Full w/ SeqCat.                   &     3.66   &    3.94    \\
LoRA w/ EmbedAdd.\ (Ours)         &   6.11     &  6.29      \\
LoRA w/ SeqCat.\ (Ours)           &     7.05   &     7.47   \\
\bottomrule
\end{tabular}
}
\vspace{-4mm}
\end{table}

\mysection{Using I2I datasets}  
Since generating V2V editing datasets is much more expensive than creating I2I editing datasets, and because strong I2I models and high-quality open-source datasets already exist, we argue that V2V editing models should also be trained on I2I data. A simple approach is to treat image editing samples as single-frame videos, but this lacks motion. To narrow the domain gap, we apply a consistent sequence of affine transformations to the input and edited images to simulate pseudo V2V editing pairs.  
We perform this ablation study by sampling an equal number of I2I and V2V editing pairs and train different models for different dataset compositions. 
We test performance on a benchmark containing 100 videos from edit types supported by both datasets. 
As shown in Table~\ref{tab:i2i_dataset_ablation}, adding affine transformations improves downstream V2V editing performance, and jointly training on I2I datasets outperforms training solely on V2V datasets.

\begin{table}[t]
\centering
\caption{Ablation study on using I2I data. \textit{Single Image} treats I2I pairs as single-frame videos, while \textit{Affine Image} applies affine transformations to I2I data to create V2V pairs. } 
\vspace{-3mm}
\label{tab:i2i_dataset_ablation}
\resizebox{0.8\columnwidth}{!}{%
\begin{tabular}{@{}ccc cc@{}}
\toprule
\multicolumn{3}{c}{\textbf{Training Datasets}} & \textbf{VLM Eval.} & \textbf{Video Quality} \\
\cmidrule(r){1-3} \cmidrule(lr){4-4} \cmidrule(l){5-5}
\textbf{Single Image} & \textbf{Affine Image} & \textbf{Video Edit} & \textbf{Edit Quality} & \textbf{Pick Score} \\
\midrule
\cmark & \xmark & \xmark     & 5.52 & 19.49 \\
\cmark & \cmark & \xmark     & 6.24 & 19.67 \\
\xmark & \xmark & \cmark     & \cellcolor{peach!50}6.69 & \cellcolor{peach!50}19.90 \\
\cmark & \cmark & \cmark     & \cellcolor{mintgreen!100}\textbf{6.86} & \cellcolor{mintgreen!100}19.94 \\
\bottomrule
\end{tabular}%
}
\end{table}
\vspace{-1mm}
\textbf{}

\mysection{Using V2V and dense-captioned datasets} 
To examine the effect of data curated by our pipeline, we train different models on various V2V datasets using the same number of training steps. We then evaluate performance on a mini benchmark with 10 videos for each edit type. 
As shown in Table~\ref{tab:v2v_dataset_ablation}, training on each of our proposed V2V datasets significantly improves performance on certain edit types, which validates the effectiveness of our proposed data pipeline. The only exception is our \emph{Human Animate} dataset, which is outperformed by our \emph{Actor Transmutation} dataset according to the VLM score. Although both tasks are similar, the latter includes more diverse subjects. However, the Human Animate dataset remains useful for preserving human identities across edits and maintaining consistent facial expressions. 
\begin{table}[t]
\centering
\caption{ Ablation study on the effectiveness of our proposed video datasets using the VLM score as the metric. }
\vspace{-3mm}
\label{tab:v2v_dataset_ablation}
\resizebox{\columnwidth}{!}{%
\begin{tabular}{lccccccc}
\toprule
  & \multicolumn{7}{c}{\textbf{Training on Datasets}} \\
\cmidrule(l){2-8}
\textbf{Edit types} 
& \makecell[c]{\scriptsize Señorita-2M~\cite{zi2025senorita}} 
& \makecell[c]{\scriptsize Stylization} 
& \makecell[c]{\scriptsize Human Animate} 
& \makecell[c]{\scriptsize Controllable Video} 
& \makecell[c]{\scriptsize Flow Edit} 
& \makecell[c]{\scriptsize Inpainting} 
& \makecell[c]{\scriptsize Dense Caption} \\
\midrule
Stylization    & 4.97 & \cellcolor{mintgreen!100}\textbf{7.97} & 5.33 & 5.30   & \cellcolor{peach!50}5.77 & 4.63 & 5.20  \\
Animation      & 3.88 & 3.65 & \cellcolor{peach!50}7.20 & 3.80  & \cellcolor{mintgreen!100}\textbf{7.48} & 3.13 & 4.18  \\
Change object  & 3.33 & 2.40 & \cellcolor{peach!50}4.43 & 3.20  & \cellcolor{mintgreen!100}\textbf{5.13} & 3.13 & 4.27  \\
Control video  & \cellcolor{peach!50}5.47 & 4.42 & 5.20 & \cellcolor{mintgreen!100}\textbf{6.13} & 3.46 & 4.00 &5.04 \\
Actor transmutation & 4.37 & 2.53 & \cellcolor{peach!50}6.23 & 3.90 & \cellcolor{mintgreen!100}\textbf{8.30} & 2.83 & 5.00\\
Edit w/ Mask & \cellcolor{peach!50}3.40 & 2.73 & 3.17 & 3.03 & 2.43 & \cellcolor{mintgreen!100}\textbf{4.63} & 1.10 \\
Change human action   & 4.97 & 4.50 & 5.03 & 3.53 & \cellcolor{peach!50}6.50 & 2.50 & \cellcolor{mintgreen!100}\textbf{6.87} \\
\bottomrule
\end{tabular}%
}
\vspace{-4mm}
\end{table}

\begin{figure}
    \centering
    \includegraphics[width=\columnwidth]{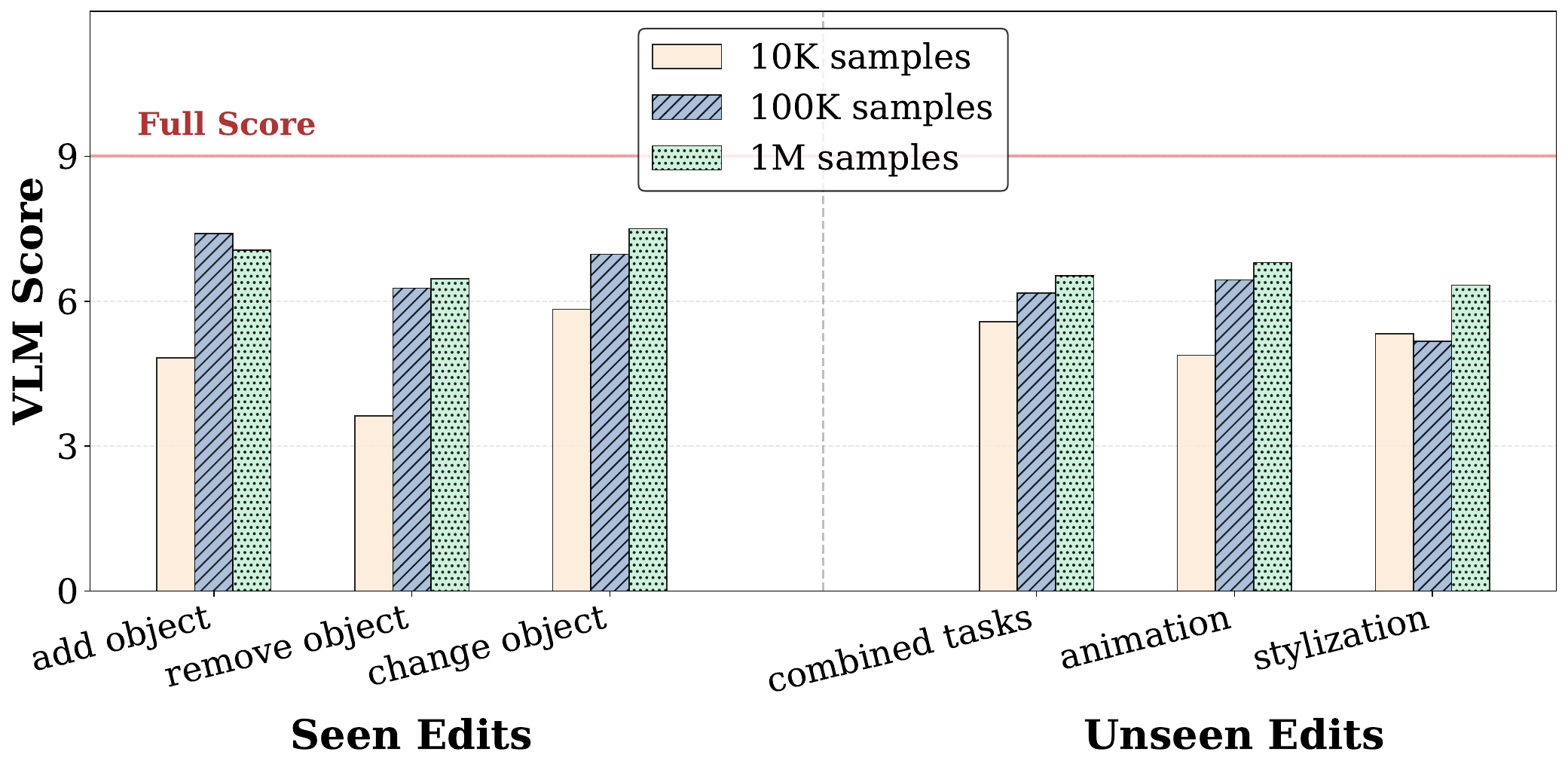}
    \vspace{-7mm}
    \caption{ Ablation study on training data size and generalization.}
    \label{fig:dataset_size}
    \vspace{-3mm}
\end{figure}

\mysection{Impact of dataset size and generalization to unseen edits}  
We split a subset of our V2V training data to include three edit types only and ablate on training data size with $10K$, $100K$, and $1M$ samples. As training data size increases, Figure~\ref{fig:dataset_size} shows performance improves accordingly for both seen and unseen edits. We observe that training with only 10K examples already yields fair performance. Moreover, editing capability on seen tasks consistently enhances performance on unseen edit categories, validating that the inherent edit ability of a pretrained T2V model can be unlocked with our efficient tuning.

\section{Conclusion}
\label{sec:conclusion}

We introduced \emph{EasyV2V}, a lightweight, instruction-based video editor that achieves state-of-the-
performance.
First, we propose robust data engines that curate diverse training data, notably by ``lifting'' static image edits into dynamic pseudo-video pairs using shared affine motion, supplemented with video continuation supervision from text2video data.
Second, we propose a parameter-efficient architecture which effectively conditions on the source video, enabling unified spatiotemporal control from a single mask with optional reference images.
This combination of a novel data strategy and a minimal-tuning architecture provides a strong and effective recipe for high-quality video editing with extensive controllability.

\noindent\textbf{Limitations.} 
While EasyV2V is robust on a wide range of edit types, common to other concurrent diffusion-based video models, inference takes around one minute and precludes its use in real-time applications.
This framework can be naturally extended to support more advanced control abilities. For example, adding geometric cinematographic camera pose controls is a compelling direction to further broaden our creative applications.

\maketitlesupplementary

\renewcommand{\thesection}{\Alph{section}}
\renewcommand{\thetable}{\Roman{table}}
\renewcommand{\thefigure}{\Roman{figure}}

\setcounter{section}{0}
\setcounter{table}{0}
\setcounter{figure}{0}

\begin{table*}[t]
\caption{Category-wise image-editing performance on ImgEdit Bench~\cite{ye2025imgeditbench}. Scores are derived from a vision–language model (VLM) based on \textit{Prompt Compliance}, \textit{Visual Naturalness/Seamlessness}, and \textit{Physical/Detail Coherence} (higher is better). \textbf{Impressively, EasyV2V surpasses all baselines across most categories and approaches the performance of leading commercial systems}, despite not being specifically designed for image editing.}
\label{tab:image_edit}
\centering
\small
\setlength{\tabcolsep}{4.5pt}
\begin{tabular}{@{}lcccccccccc@{}}
\toprule
\textbf{Method} &
\textbf{Add} & \textbf{Adjust} & \textbf{Extract} & \textbf{Replace} & \textbf{Remove} &
\textbf{Background} & \textbf{Style} & \textbf{Hybrid} & \textbf{Action} &
\textbf{Overall}\,$\uparrow$ \\
\midrule
MagicBrush      & 2.84 & 1.58 & 1.51 & 1.97 & 1.58 & 1.75 & 2.38 & 1.62 & 1.22 & 1.83 \\
Instruct-P2P    & 2.45 & 1.83 & 1.44 & 2.01 & 1.50 & 1.44 & 3.55 & 1.20 & 1.46 & 1.88 \\
AnyEdit         & 3.18 & 2.95 & 1.88 & 2.47 & 2.23 & 2.24 & 2.85 & 1.56 & 2.65 & 2.45 \\
UltraEdit       & 3.44 & 2.81 & \cellcolor{peach!50} 2.13 & 2.96 & 1.45 & 2.83 & 3.76 & 1.91 & 2.98 & 2.70 \\
ICEdit          & 3.58 & 3.39 & 1.73 & 3.15 & 2.93 & 3.08 & 3.84 & 2.04 & 3.68 & 3.05 \\
Step1X-Edit     & \cellcolor{peach!50} 3.88 & 3.14 & 1.76 & 3.40 & 2.41 & 3.16 & \cellcolor{peach!50} 4.63 & 2.64 & 2.52 & 3.06 \\
UniWorld-V1     & 3.82 & 3.64 & \cellcolor{mintgreen!100} 2.27 & 3.47 & \cellcolor{peach!50} 3.24 & 2.99 & 4.21 & 2.96 & 2.74 & 3.26 \\
BAGEL           & 3.81 & 3.59 & 1.58 & 3.85 & 3.16 & 3.39 & 4.51 & 2.67 & 4.25 & 3.42 \\
OmniGen2        & 3.57 & 3.06 & 1.77 & 3.74 & 3.20 & 3.57 & \cellcolor{mintgreen!100} 4.81 & 2.52 & \cellcolor{mintgreen!100} 4.68 & 3.44 \\
Kontext-dev     & 3.83 & \cellcolor{peach!50} 3.65 & \cellcolor{mintgreen!100} 2.27 & \cellcolor{mintgreen!100} 4.45 & 3.17 & \cellcolor{peach!50} 3.98 & 4.55 & \cellcolor{peach!50} 3.35 & \cellcolor{peach!50} 4.29 & \cellcolor{peach!50} 3.71 \\
\textbf{EasyV2V (Ours)} & \cellcolor{mintgreen!100} 4.46 & \cellcolor{mintgreen!100} 4.18 & 1.80 & \cellcolor{peach!50} 3.86 & \cellcolor{mintgreen!100} 3.70 & \cellcolor{mintgreen!100} 4.33 & 4.57 & \cellcolor{mintgreen!100} 4.04 & \cellcolor{mintgreen!100} 4.68 & \cellcolor{mintgreen!100} 3.96 \\

\hdashline
\multicolumn{11}{l}{\textit{Closed-Source Commercial Models and Concurrent works}} \\
\textcolor{gray!60}{EditVerse} & \textcolor{gray!60}{3.81} & \textcolor{gray!60}{3.62} & \textcolor{gray!60}{1.44} & \textcolor{gray!60}{3.95} & \textcolor{gray!60}{3.14} & \textcolor{gray!60}{3.58} & \textcolor{gray!60}{4.71} & \textcolor{gray!60}{2.72} & \textcolor{gray!60}{3.80} & \textcolor{gray!60}{3.42} \\
\textcolor{gray!60}{Ovis-U1}         & \textcolor{gray!60}{3.99} & \textcolor{gray!60}{3.73} & \textcolor{gray!60}{2.66} & \textcolor{gray!60}{4.38} & \textcolor{gray!60}{4.15} & \textcolor{gray!60}{4.05} & \textcolor{gray!60}{4.86} & \textcolor{gray!60}{3.43} & \textcolor{gray!60}{4.68} & \textcolor{gray!60}{3.97} \\
\textcolor{gray!60}{GPT-4o-Image}    & \textcolor{gray!60}{4.61} & \textcolor{gray!60}{4.33} & \textcolor{gray!60}{2.90} & \textcolor{gray!60}{4.35} & \textcolor{gray!60}{3.66} & \textcolor{gray!60}{4.57} & \textcolor{gray!60}{4.93} & \textcolor{gray!60}{3.96} & \textcolor{gray!60}{4.89} & \textcolor{gray!60}{4.20} \\

\bottomrule
\end{tabular}
\end{table*}

\section{Performance on image editing}

During training, we occasionally sampled from image-editing datasets and treated each image–edit pair as a single-frame video, rather than exclusively applying affine transformations to synthesize pseudo-videos.

As shown in Table~\ref{tab:image_edit}, we evaluated EasyV2V on a recent image-editing benchmark~\cite{ye2025imgeditbench}, interpreting each image as a single-frame video with a resolution of $1\times832\times480$. \textit{Impressively, although EasyV2V was not specifically designed for image editing, it surpassed all baselines on nearly all subtasks and achieved performance comparable to leading closed-source and commercial systems.}

\begin{figure*}
    \centering
    \includegraphics[width=\linewidth]{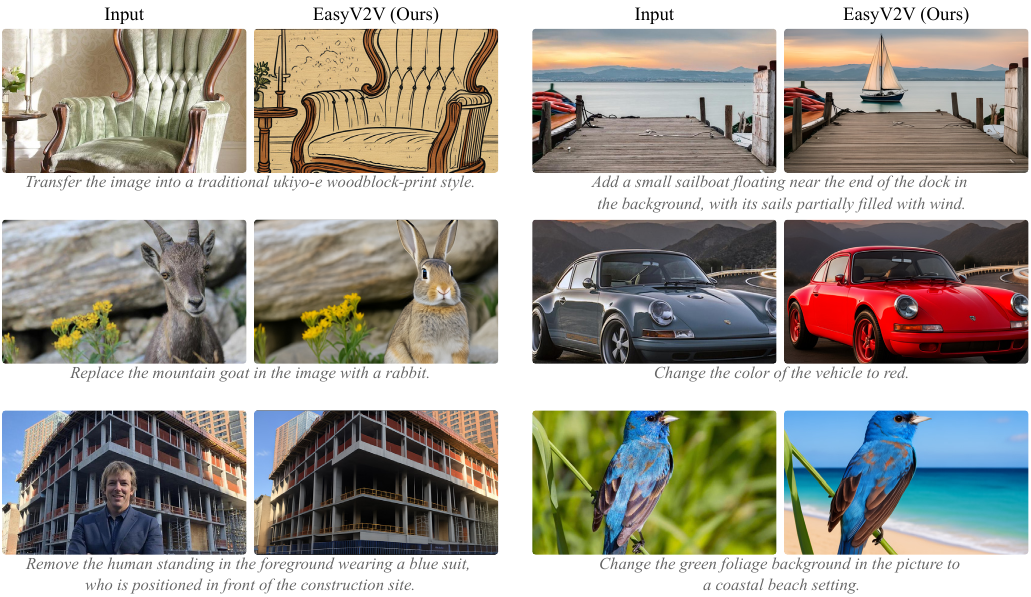}
    \caption{Qualitative image-editing results from our video-editing model, which treats images as single-frame videos and achieves state-of-the-art performance, showing that video editing aids image editing.}
    \label{fig:image_editing_qualitative}
\end{figure*}

Notably, EasyV2V outperformed EditVerse by a margin of \textbf{0.54}. The model also demonstrated strong results on both \textbf{action} and \textbf{hybrid} editing tasks, underscoring the effectiveness of our unified data pipeline that jointly leverages image-editing datasets and video–caption datasets featuring human actions. Figure~\ref{fig:image_editing_qualitative} presents representative qualitative results, with additional examples provided in our supplementary materials.

\section{Model configuration}
\subsection{Mask conditioning} 
 
\begin{figure*}[!ht]
\centering
\begin{subfigure}[t]{0.30\textwidth}
  \centering
  \adjustbox{valign=t, min totalheight=0.2\textheight}{%
    \includegraphics[width=\linewidth]{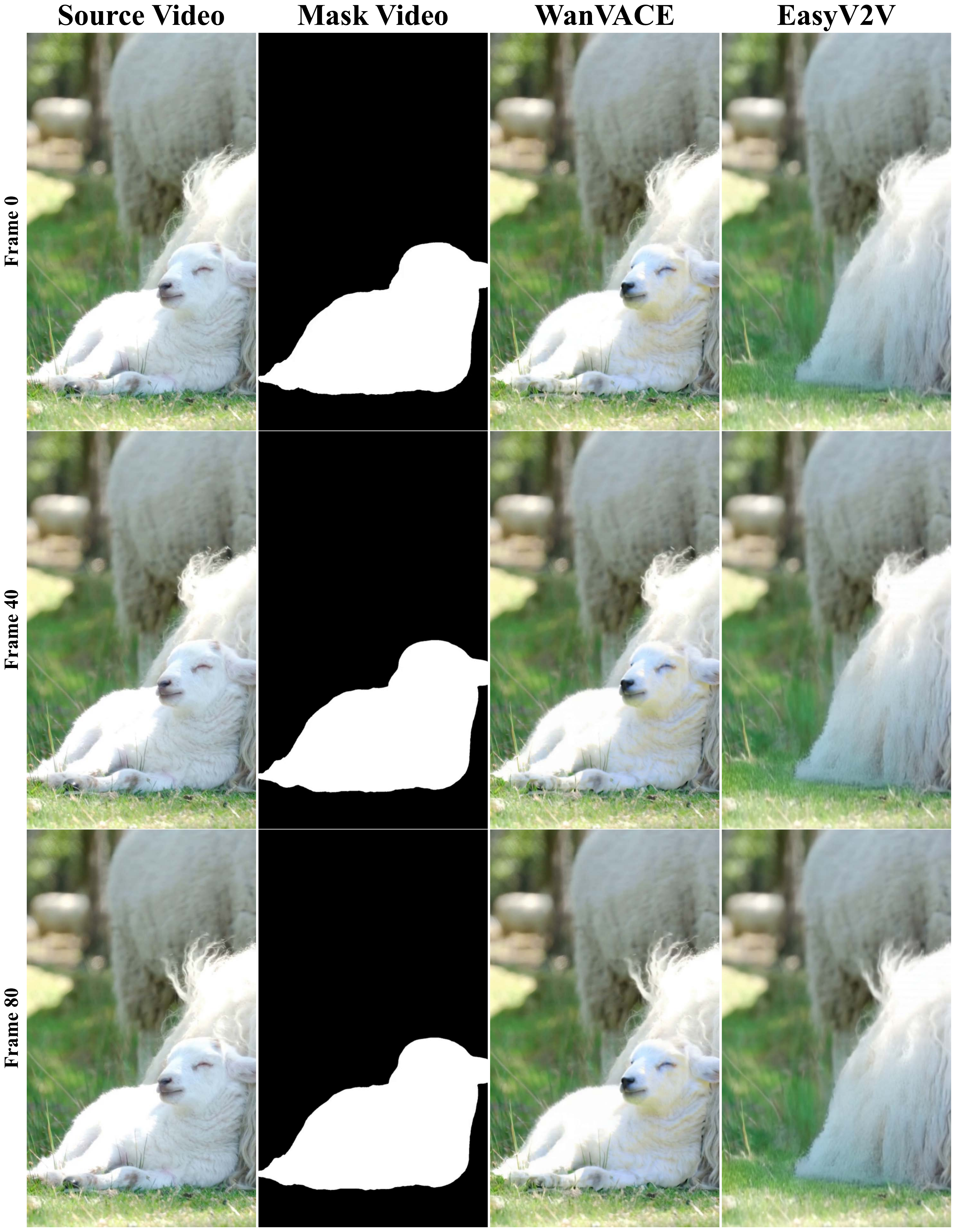}}
  \caption{Edit with spatial mask. Edit prompt: \textit{``remove the lamb.''}}
\end{subfigure}\hfill
\begin{subfigure}[t]{0.70\textwidth}
  \centering
  \adjustbox{valign=t, min totalheight=0.2\textheight}{%
    \includegraphics[width=\linewidth]{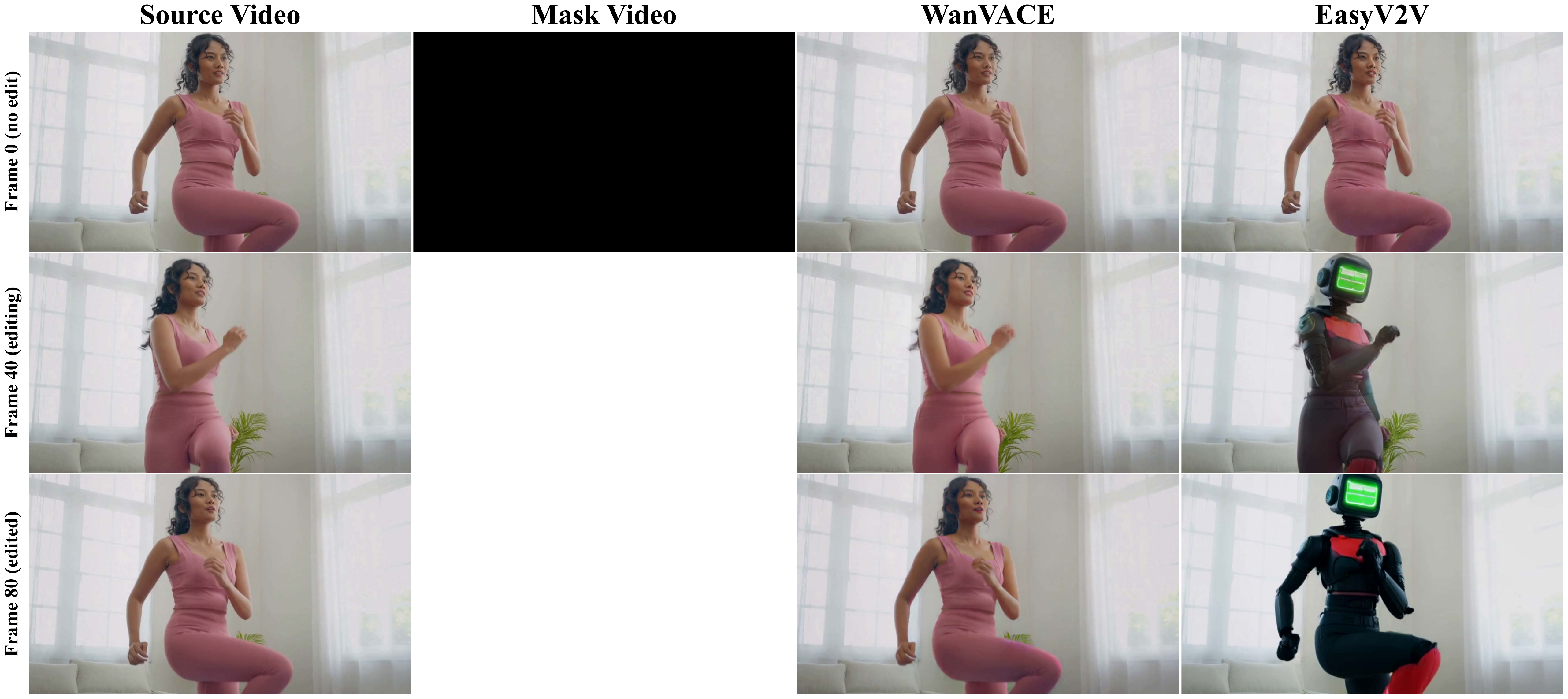}}
  \caption{Edit with temporal mask. Edit prompt: \textit{``change the woman to a robot.''}}
\end{subfigure}
\caption{Comparison of mask-guided editing performance with WanVACE using samples from EditVerseBench~\cite{editverse2025}.}
\label{fig:mask_vace}
\end{figure*}

We compare our mask-conditioned editing performance with WanVACE~\cite{vace}.
Note that as WanVACE is not an instruction-based editing model, we restrict our comparison to the edit types it supports via edit masks: \emph{inpainting}, \emph{object removal}, and \emph{object replacement}.
We evaluate two types of edit masks: pixel-wise spatial masks (indicating the edit region per frame) and frame-wise temporal masks (indicating which frames should be modified).
While WanVACE utilizes an additional branch with complex context activation and injection for mask conditioning, we demonstrate that encoding masks using the video VAE is simple yet effective, particularly for temporal control.

We investigate different mask conditioning strategies for EasyV2V:
\begin{itemize}
    \item Video VAE, $Z_{\text{msk}} + Z_{\text{src}}$: After encoding the mask video into mask latent $Z_{\text{msk}}$ using the video VAE, we perform token addition to inject the mask condition into the source video latent. This is the default strategy for EasyV2V in the main paper.
    \item Video VAE, $Z_{\text{msk}} + Z_{\text{tgt}}$: After encoding the mask video into mask latent $Z_{\text{msk}}$ using the video VAE, we perform token addition to inject the mask condition into the noisy target video latent.
    \item Downsample, $Z_{\text{msk}} + Z_{\text{src}}$: We apply spatial and temporal average pooling to downsample the mask video from input resolution to latent resolution, then perform token addition to inject the mask condition into the source video latent.
    \item Video VAE, $\texttt{Seq\_Cat}(Z_{\text{src}}, Z_{\text{tgt}}, Z_{\text{msk}})$: After encoding the mask video into mask latent $Z_{\text{msk}}$ using the video VAE, we perform sequence concatenation for all condition signals, including $Z_{\text{src}}, Z_{\text{tgt}}, Z_{\text{msk}}$, and $Z_{\text{ref}}$.
\end{itemize}

To evaluate spatial mask editing, we adopt the VLM prompt from the EditVerse evaluation protocol.
To evaluate temporal mask editing, where the edit occurs after a certain timestamp, we modify the VLM prompt to incorporate edit timestamp awareness. Please refer to the last page for our complete VLM prompts.

As shown in Table~\ref{tab:mask_ablation}, EasyV2V achieves the best performance when using token addition to inject the mask condition into the source video latent. Both sequence concatenation and our token addition strategy outperform WanVACE on mask-conditioned video editing.

\begin{table*}[ht]
\centering
\caption{Comparison of video editing mask strategies.}
\label{tab:mask_ablation}
\resizebox{1.8\columnwidth}{!}{%
\begin{tabular}{lccc} %
\toprule
\textbf{Method} & \multicolumn{2}{c}{\textbf{VLM score} ($\uparrow$)} & \\ %
\cmidrule(lr){2-4} %
& \textbf{Pixel-wise spatial mask} & \textbf{Frame-wise temporal mask} & \textbf{Average} \\ %
\midrule
Wan VACE~\cite{vace} & 4.13 & 6.87 & 5.50 \\ %
Video VAE, $Z_{\text{msk}} + Z_{\text{tgt}}$ & 5.50 & 7.40 & 6.45 \\ %
Downsample, $Z_{\text{msk}} + Z_{\text{src}}$ & 5.80 & 5.40 & 5.60 \\ %
Video VAE, $\texttt{Seq\_Cat}(Z_{\text{src}}, Z_{\text{tgt}}, Z_{\text{msk}})$ & \cellcolor{peach!50}6.00 & \cellcolor{peach!50}7.70 & \cellcolor{peach!50}6.85 \\ %
Video VAE, $Z_{\text{msk}} + Z_{\text{src}}$ (EasyV2V) & \cellcolor{mintgreen!100}7.23 & \cellcolor{mintgreen!100}7.73 & \cellcolor{mintgreen!100}7.48 \\ 
\bottomrule
\end{tabular}
}
\end{table*}

From the qualitative comparison in Figure~\ref{fig:mask_vace}, we observe that WanVACE~\cite{vace} has limited ability to generalize to diverse edit prompts and fails to adhere to both temporal and spatial masks in many cases compared to EasyV2V.

\subsection{LoRA Rank}

We employ LoRA~\cite{hu2022lora} to mitigate divergence and study the effect of LoRA rank. 
We train the model for 20K steps to ablate the impact of LoRA rank on the EditVerse benchmark. 
We observe that performance improves as we increase the rank. 
We adopt a LoRA rank of 256, as performance begins to saturate between ranks 128 and 256. 
Table~\ref{tab:lora_rank_ablation} reports the results for ranks 64, 128, and 256. 
Although the model with rank 256 performs best on most metrics, the gap between ranks 128 and 256 is marginal, and rank 64 is only slightly inferior. 
This supports our hypothesis that pre-trained video models serve as strong priors for video editing, and that a low-rank update is sufficient for robust performance.

\begin{table}[t]
  \centering
  \caption{Ablation study on LoRA rank and the use of the reference image. }
  \label{tab:lora_rank_ablation} 
  \resizebox{\columnwidth}{!}{%
  \begin{tabular}{lcccccc}
    \toprule
    & \multicolumn{2}{c}{Rank 64} & \multicolumn{2}{c}{Rank 128} & \multicolumn{2}{c}{Rank 256} \\
    \cmidrule(lr){2-3} \cmidrule(lr){4-5} \cmidrule(lr){6-7}
    Metric & w/o Ref. & w/ Ref. & w/o Ref. & w/ Ref. & w/o Ref. & w/ Ref. \\
    \midrule
    Frame-Text Alignment      & 24.80 & \cellcolor{peach!50} 27.27 & 24.99 & \cellcolor{mintgreen!100} \textbf{27.67} & 25.32 & \cellcolor{mintgreen!100} \textbf{27.67} \\
    Video-Text Alignment      & 21.31 & 24.36 & 21.45 & \cellcolor{mintgreen!100} \textbf{24.73} & 21.90 & \cellcolor{peach!50} 24.60 \\
    PickScore Video Quality   & 19.42 & 20.20 & 19.52 & \cellcolor{peach!50} 20.36 & 19.58 & \cellcolor{mintgreen!100} \textbf{20.37} \\
    \textbf{VLM Quality Score}    & 6.17  & 7.02  & 6.20  & \cellcolor{peach!50} 7.12  & 6.48  & \cellcolor{mintgreen!100} \textbf{7.22}  \\
    \bottomrule
  \end{tabular}%
  }
\end{table}

\subsection{Time and memory profiling}

We profile the efficiency of EasyV2V on a single NVIDIA H100 GPU for training and inference under different strategies:
(1) full model fine-tuning with sequence concatenation of source and target latents,
(2) full model fine-tuning with token addition of source and target latents,
(3) LoRA fine-tuning with sequence concatenation of source and target latents, and
(4) LoRA fine-tuning with token addition of source and target latents.
FlashAttention is enabled during both training and inference.
A complete comparison is provided in Table~\ref{tab:arch_efficiency}.

\begin{table}[t]
\caption{Comparison of training and inference costs across different tuning strategies.}
\label{tab:arch_efficiency}
\centering
\small
\setlength{\tabcolsep}{6pt}
\resizebox{\columnwidth}{!}{%
\begin{tabular}{@{}lcccc@{}}
\toprule
\textbf{Metric} & \textbf{Full Model w/ SeqCat.} & \textbf{Full Model w/ EmbedAdd.} & \textbf{LoRA w/ SeqCat.} & \textbf{LoRA w/ EmbedAdd.} \\
\midrule
New Params                 & 5B     & 5B     & 0.64B  & 0.64B  \\
Train (s / batch)          & 5.63   & 4.60   & 5.70   & 4.54   \\
Train VRAM                 & 62\,GB & 62\,GB & 37\,GB & 32\,GB \\
Inference (s / sample)     & 67.71  & 30.41  & 69.42  & 30.11  \\
\bottomrule
\end{tabular}
}
\end{table}

\section{Additional details on data pipelines}

\noindent\textbf{Human animation.}
We use the 14B-parameter pretrained Wan Animate~\cite{wananimate2025} model as the expert editor, following its preprocessing for face crops and poses. For first-frame editing, we apply Flux Kontext Dev~\cite{labs2025flux1kontextflowmatching}. Edit prompts are generated by ChatGPT~\cite{openai2024gpt4ocard} using 150 prompts created from the following instruction:

\begin{vlmprompt}
You are a helpful assistant to help with the generation of video editing prompts, to edit videos of people. Below are samples of the prompts we're interested it:

# Fantasy & Mythical Creatures
- "Make the person look like an elf"
- "Make the person look like a goblin"

# Professions & Archetypes
- "Make the person look like a knight"
- "Make the person look like a samurai"

# Horror & Dark Styles
- "Make the person look like a zombie"
- "Make the person look like a vampire"

# Animals & Hybrids
- "Make the person look like a lion"
- "Make the person look like a tiger"

# Sci-Fi & Futuristic
- "Make the person look like a robot"
- "Make the person look like a cyborg"

# Stylized & Surreal
- "Make the person look like a stained-glass figure"
- "Make the person look like a chalk drawing"

# Accessories
- "Give the person a pair of sunglasses"
- "Give the person a hat"

Now, generate several prompts per category mentioned above. 
\end{vlmprompt}

\noindent\textbf{Actor transmutation.}

We adapt FlowEdit~\cite{kulikov2024flowedit}, originally designed for image editing, on Wan 2.1 14B~\cite{wan2025wanopenadvancedlargescale} for video editing. As described in the main paper, our prompt assets include three object categories: bipedals (e.g., clown, pirate, ninja, samurai, humanoid robot), quadrupeds (e.g., dog, cat, lion, cheetah, sheep), and avians (e.g., pigeon, duck, parrot, eagle, owl). We provide five examples per category to ChatGPT to generate extended lists. We also compile lists of scenes (e.g., jungle, mountain, beach, street, bedroom) and actions (e.g., walking, running, jumping, dancing).

\noindent\textbf{Video stylization.}

Similar to the prompts for our human animate dataset, we use ChatGPT to generate prompts for video stylization:

\begin{vlmprompt}
You are a helpful assistant to help with the generation of video stylization prompts, to edit and stylize in the wild videos. Below are samples of the prompts we're interested it:

# Distinct media / aesthetics
- "In the style of a watercolor painting"
- "In the style of a digital painting"

# Comic / cartoon & animation houses
- "In the style of a manga"
- "In the style of a cartoon"

# Art movements
- "In the style of Brutalism"
- "In the style of Impressionism"

# Lighting
- "Captured in the golden hour"
- "Captured bathed in neon lights"

# Cinematic & photographic framing styles
- "Shot on 35 mm film grain"
- "Shot in black-and-white film noir style"

# Global traditional arts & patterns
- "In the style of a Roman floor mosaic"
- "In the style of Persian miniature painting"

# Color palette & tonal approaches
- "Rendered in soft pastel hues"
- "Rendered in muted earth tones"

Now, generate several prompts per category mentioned above. 
\end{vlmprompt}

We end up generating 350 different stylization prompts. 

\noindent\textbf{Controllable video generation.}

We curate a 15K-sample dataset from in-the-wild videos through manual filtering, then apply a range of model-free and model-based video-to-video transformations to build a paired dataset for training our video editing model. These transformations include human pose estimation (DWPose), Canny and HED edge detection, RAFT large optical flow, random black rectangle masks (inpainting), random black borders (outpainting), depth prediction with Depth Anything V2, grayscale conversion, Gaussian blur, color negation, saturation/contrast/brightness adjustments, pixelation, wave warping, posterization, Gaussian noise, and color overlays.

\noindent\textbf{Dense-captioned text-to-video data}

We apply strict filtering criteria to ensure high-quality training pairs from dense-captioned datasets.
First, we require videos to have at least 162 frames after downsampling to 15 fps (enabling 81 frames for both source and target clips).
Second, we filter temporal segments based on: (i) start time must allow sufficient preceding frames ($t_i \geq 81/\text{fps}_{\text{downsample}}$), (ii) segment duration must exceed 2 seconds to ensure meaningful actions, and (iii) no scene cuts within the segment interval.
For instruction generation, we flatten all temporal segments from multiple videos into batches and process them simultaneously with Qwen-3-4B~\cite{qwen3}, discarding segments where the LLM returns empty strings (indicating unsuitable actions for video editing conversion).
The LLM prompt instructs the model to convert action descriptions into imperative instructions starting with verbs like ``make,'' ``let,'' or ``have,'' while preserving all key details.

\noindent\textbf{Image-to-image data}

We employ a multi-stage pipeline to generate high-quality instructional I2I pairs from image captions at scale.
Given an image caption, an LLM (Qwen3-4B-Instruct~\cite{qwen3}) generates up to five diverse edit instructions spanning canonical edit types: \texttt{add}, \texttt{remove}, \texttt{replace}, \texttt{change\_global}, \texttt{change\_local}, \texttt{change\_color}, \texttt{transform\_global}, \texttt{transform\_local}, \texttt{text}, and \texttt{other}.
Each instruction is then executed using instruction-following image editors (Qwen-Image-Edit~\cite{qwenimage2025} , or Flux-Kontext~\cite{labs2025flux1kontextflowmatching}), producing candidate edit pairs.
To ensure quality, we apply VLM-based filter with Gemma-3-27B~\cite{gemma3}.

For I2I-to-V2V conversion, we generate smooth affine camera trajectories by sampling random target poses with bounded parameters: rotation angles in $\left[-15^\circ, 15^\circ\right]$, zoom factors in $[0.66, 1.0]$ (avoiding excessive zoom-out), and translation offsets within $\pm 33\%$ of frame dimensions.
These parameters are linearly interpolated across frames and constrained via linear programming to ensure the transformed bounding box remains fully within the frame boundaries throughout the trajectory.
The same trajectory is applied to both source and target images using perspective transforms, creating temporally coherent pseudo-videos with motion cues (zoom, pan, rotation) while preserving the edit signal.
With 50\% probability, trajectories are reversed to balance zoom-in and zoom-out motions, providing diverse camera movement patterns that enhance the model's robustness to dynamic viewpoints during training.

\section{Impact of classifier free guidance}

We perform an ablation study on the impact of classifier-free guidance (CFG)~\cite{CFG} during inference.

\paragraph{CFG Implementation.}
Following the standard CFG formulation, we guide the denoising process by interpolating between conditional and unconditional predictions. Given the predicted noise $\epsilon_\theta$ from our diffusion model, the CFG-guided prediction is computed as:
\begin{equation}
\hat{\epsilon}_\theta = \epsilon_\theta(\emptyset) + s \cdot (\epsilon_\theta(c) - \epsilon_\theta(\emptyset))
\end{equation}
where $c$ represents the conditioning signal, $\emptyset$ denotes the unconditional (null) condition, and $s$ is the guidance scale. This can be reformulated as:
\begin{equation}
\hat{\epsilon}_\theta = (1-s) \cdot \epsilon_\theta(\emptyset) + s \cdot \epsilon_\theta(c)
\end{equation}
When $s = 1.0$, the model performs purely conditional generation, while larger values of $s$ amplify the influence of the conditioning signal.

In our framework, we support two CFG strategies depending on which conditions are used for guidance: \emph{Prompt-only CFG} and \emph{Prompt + Reference CFG}.

\noindent\textbf{Prompt-only CFG.} By default, we apply CFG only to the text prompt while keeping all visual conditions (source video and reference image) shared between conditional and unconditional branches:
\begin{equation}
\hat{\epsilon}_\theta = \epsilon_\theta(c_\text{vis}) + s \cdot (\epsilon_\theta(c_\text{vis}, c_\text{prompt}) - \epsilon_\theta(c_\text{vis}))
\end{equation}
where $c_\text{vis}$ denotes visual conditions (source video and optionally reference image), and $c_\text{prompt}$ is the edit instruction. This approach maintains consistent visual context while allowing the text prompt to guide the editing direction.

\noindent\textbf{Prompt + Reference CFG.} Alternatively, we can apply CFG to both the text prompt and reference image:
\begin{equation}
\hat{\epsilon}_\theta = \epsilon_\theta(c_\text{src}) + s \cdot (\epsilon_\theta(c_\text{src}, c_\text{ref}, c_\text{prompt}) - \epsilon_\theta(c_\text{src}))
\end{equation}
where $c_\text{src}$ is the source video (always present), $c_\text{ref}$ is the reference image, and $c_\text{prompt}$ is the text prompt. This strategy applies guidance to both the reference appearance and text instruction simultaneously. Our method is uniquely suitable for this type of guidance because the reference image is an optional input to our model.

\begin{figure}
    \centering
    \includegraphics[width=\columnwidth]{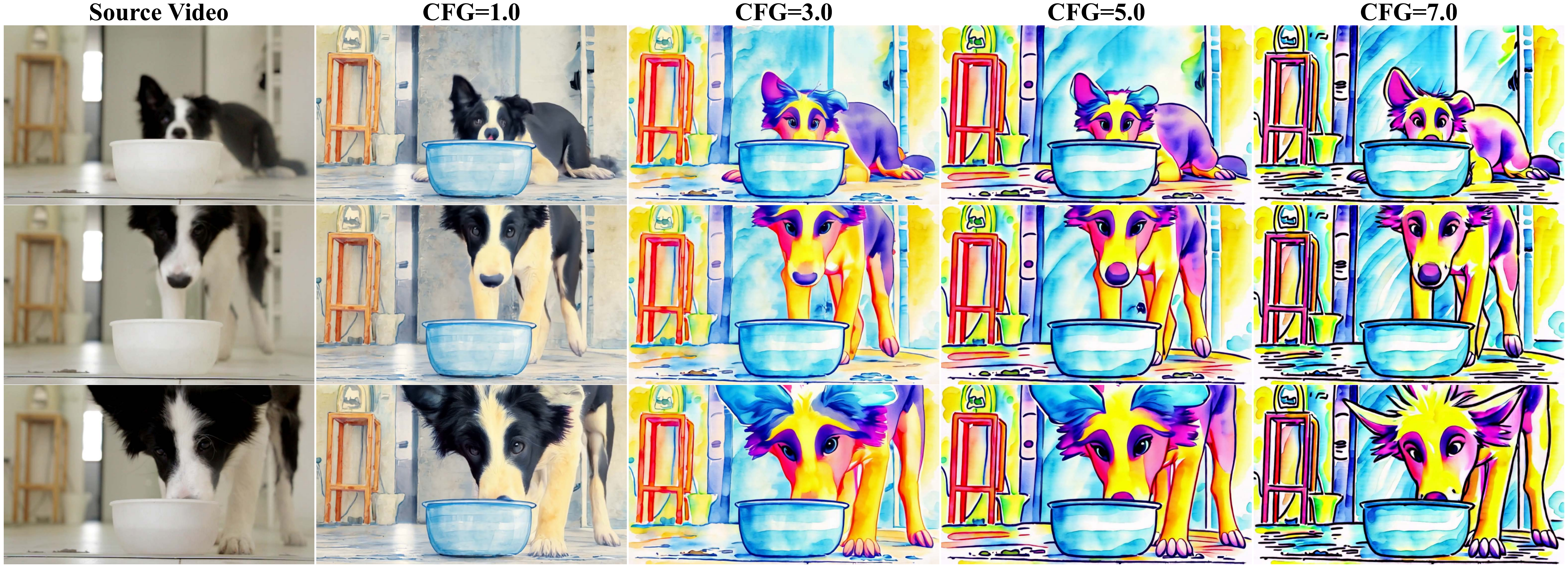}
    \caption{ 
Effect of the CFG for the edit prompt. Edit prompt: \textit{``Transform the entire visual style of the video using a hand-drawn watercolor animation effect.''}
    }
    \label{fig:cfg_prompt}
\end{figure}

\paragraph{Experimental results.}
In Table~\ref{tab:cfg_ablation}, we present results under different CFG scales when using only the edit prompt for CFG. We also provide a visualized example in Figure~\ref{fig:cfg_prompt}. As shown, higher CFG scales generally improve text alignment but may reduce temporal consistency and video quality when the scale becomes too large.

\begin{figure}
    \centering
    \includegraphics[width=\columnwidth]{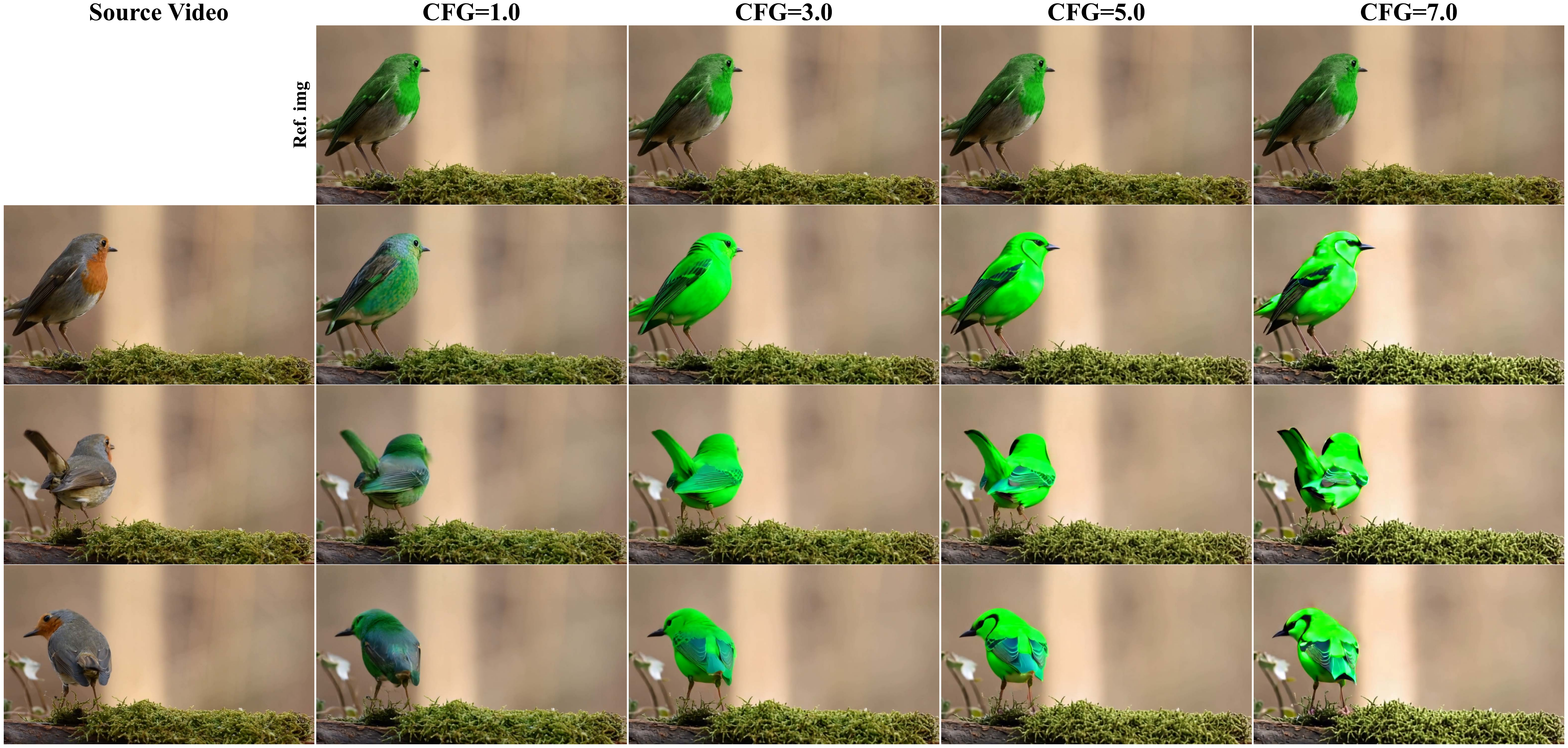}
    \caption{ 
Visual illustration of the effect of the CFG scale for the edit prompt and reference image. 
Edit prompt: \textit{``Change the bird's color to emerald green.''} 
The top row shows the reference image generated by Qwen-Image-Edit.
    }
    \label{fig:cfg_ref}
\end{figure}

In Table~\ref{tab:cfg_ablation2}, we show results under different CFG scales when using both the edit prompt and reference image for CFG. For fair comparison, we report inference performance only when a reference image is provided. A visualized example is shown in Figure~\ref{fig:cfg_ref}. The dual-condition CFG provides stronger control over both appearance and semantic alignment but requires careful tuning of the guidance scale to balance faithfulness to conditions and generation quality.

We observe that the best performance is achieved when CFG scales are between $3.0$ and $5.0$. We adopt a moderate CFG scale of $s = 3.0$ by default with prompt-only guidance. We believe that further fine-grained CFG hyperparameter tuning could yield even better performance.

\begin{table}[t]
\caption{Effect of the CFG scale for the edit prompt.}
\label{tab:cfg_ablation}
\centering
\small
\setlength{\tabcolsep}{5pt}
\resizebox{\columnwidth}{!}{%
\begin{tabular}{@{}lrrrrrrrr@{}}
\toprule
 & \multicolumn{4}{c}{\textbf{Inference w/o Ref.}} &
   \multicolumn{4}{c}{\textbf{Inference w/ Ref.}} \\
\cmidrule(lr){2-5}\cmidrule(l){6-9} %
\textbf{CFG scale} & \textbf{1.0} & \textbf{3.0} & \textbf{5.0} & \textbf{7.0}
                   & \textbf{1.0} & \textbf{3.0} & \textbf{5.0} & \textbf{7.0} \\
\midrule
Frame-Text Alignment       & 26.65 & \cellcolor{mintgreen!100} 27.59 & \cellcolor{peach!50} 27.49 & 27.11 & 27.26 & 27.28 & 27.33 & 27.29 \\
Video-Text Alignment       & \cellcolor{mintgreen!100} 24.52 & \cellcolor{peach!50} 24.46 & 24.16 & 23.40 & 24.01 & 24.07 & 24.27 & 24.33 \\
PickScore Video Quality    & 20.05 & 20.36 & 20.23 & 20.05 & 20.57 & \cellcolor{mintgreen!100} 20.60 & \cellcolor{peach!50} 20.58 & 20.53 \\
\textbf{VLM Quality Score} & 6.79 & \cellcolor{mintgreen!100} 7.73 & \cellcolor{peach!50} 7.48 & 6.98 & 7.14 & 7.30 & 7.28 & 7.21 \\

\bottomrule
\end{tabular}%
}
\end{table}

\begin{table}[t]
\caption{Effect of the CFG scale for the edit prompt and reference image.}
\label{tab:cfg_ablation2}
\centering
\small
\setlength{\tabcolsep}{5pt}
\resizebox{0.65\columnwidth}{!}{%
\begin{tabular}{lrrrr}
\toprule
 & \multicolumn{4}{c}{\textbf{Inference w/ Ref.}} \\
\cmidrule(lr){2-5}
\textbf{CFG scale} & \textbf{1.0} & \textbf{3.0} & \textbf{5.0} & \textbf{7.0} \\
\midrule
Frame-Text Alignment       & 26.65 & \cellcolor{mintgreen!100} 27.77 & \cellcolor{peach!50} 27.49 & 27.11 \\
Video-Text Alignment       & \cellcolor{peach!50} 24.52 & \cellcolor{mintgreen!100} 24.54 & 24.16 & 23.40 \\
PickScore Video Quality    & 20.05 & \cellcolor{mintgreen!100} 20.34 & \cellcolor{peach!50} 20.23 & 20.05 \\
\textbf{VLM Quality Score} & 6.79 & \cellcolor{mintgreen!100} 7.69 & \cellcolor{peach!50} 7.47 & 6.98 \\

\bottomrule
\end{tabular}
}
\end{table}

\section{User study}

\begin{figure*}
    \centering
    \includegraphics[width=1.8\columnwidth]{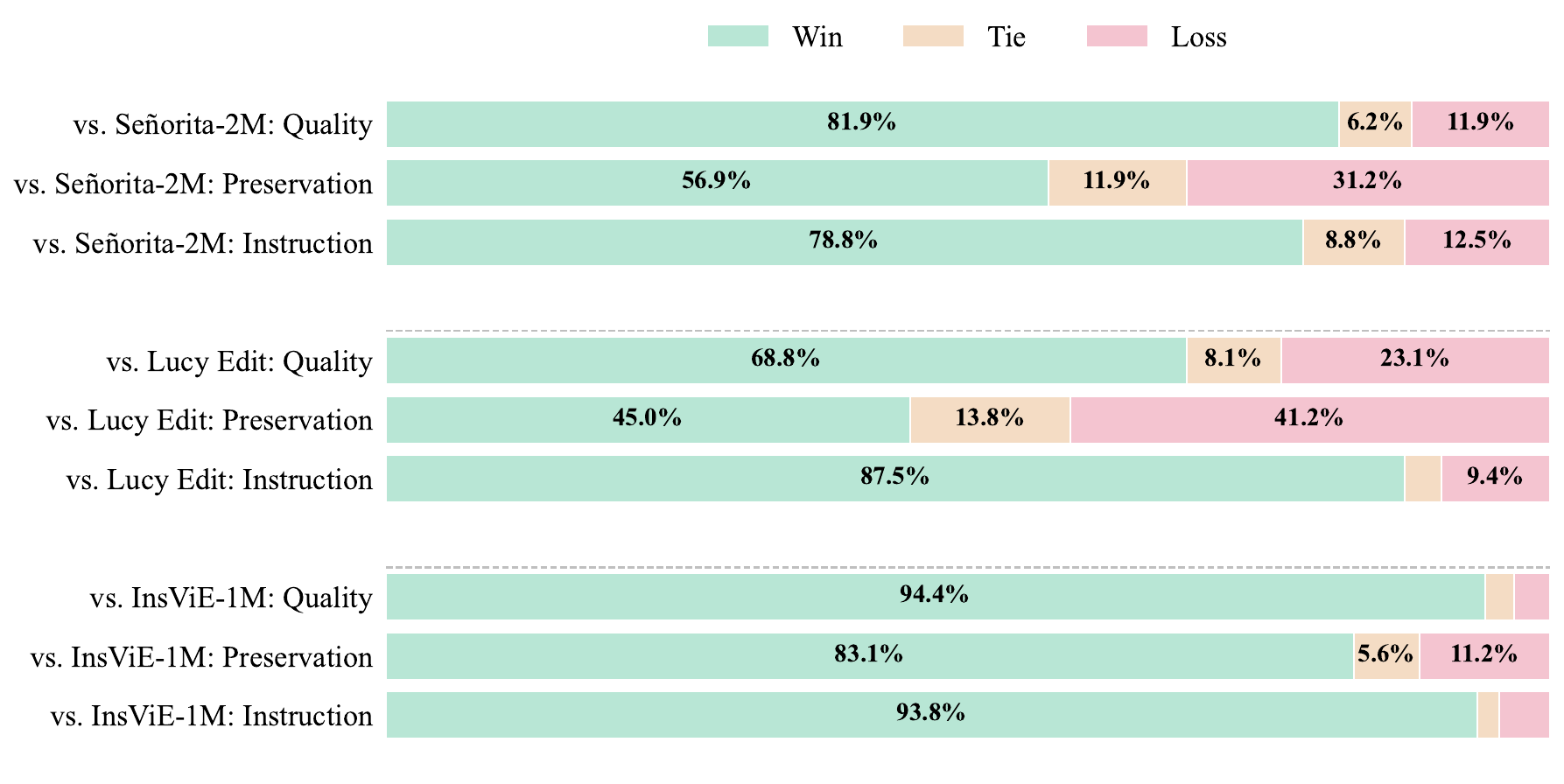}
    \caption{ Results of the user study. EasyV2V is the most preferred method across all evaluation criteria.
 }
    \label{fig:user_study}
\end{figure*}

We construct a custom benchmark comprising 160 horizontal and vertical videos spanning 18 distinct edit types, including \textit{actor transmutation, add effect, add object, animation, change action, change background, change color, change material, change object, change weather, complex edit, control video, edit with mask, freeform, local stylization, global stylization, remove object, and sim2real}.
We conduct a user study in which participants select the superior sample between outputs generated by two different methods, evaluating them along three dimensions: Instruction Alignment (adherence to the text prompt), Preservation of Unedited Regions (temporal consistency in unchanged areas), and Video Quality (overall visual fidelity).
As shown in Figure~\ref{fig:user_study}, EasyV2V outperforms all other methods across all three dimensions.

\section{Additional visualization}

\begin{figure}[!htb]
    \centering
    \includegraphics[width=\columnwidth]{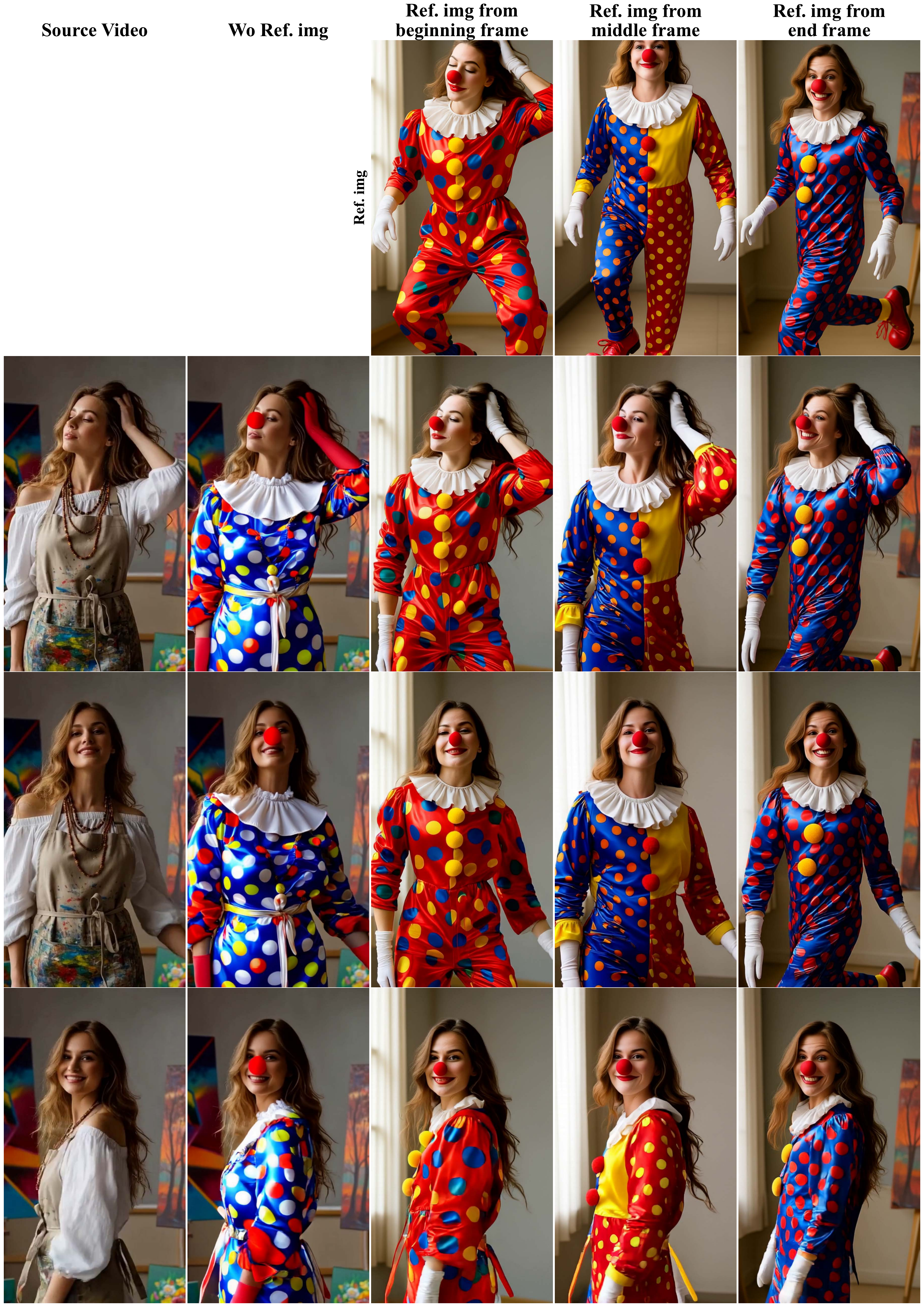}
    \caption{
Robustness of EasyV2V to the choice of reference image. Edit prompt: \textit{``Change the apron and blouse to a classic clown costume.''} 
The top row shows reference images generated by Qwen-Image-Edit based on different frames from the source video, followed by videos generated by EasyV2V conditioned on these reference images. 
The source video is taken from Lucy Edit's website~\cite{lucyedit2024}.
    }
    \label{fig:ref_robustness}
\end{figure}

\subsection{Robustness to Reference Image}

We provide visualization results of EasyV2V based on the choice of reference image.
By default, we derive the reference image by applying the image editing model to the first frame of the source video.
We also present results using the first, middle, or last frame of the source video as the basis for the reference.
As shown in Figure~\ref{fig:ref_robustness}, our model is robust to the choice of reference image, indicating that EasyV2V effectively captures the identity of the reference image for video editing.
Moreover, EasyV2V can rectify inconsistent zoom-in effects and human pose misalignments introduced by the image editing model.
Notably, without an external reference image from the image editing model, EasyV2V achieves even better consistency with the source video; for instance, the background remains well preserved.

\subsection{Comparison on Human Animate and Flow Edit Datasets}

\begin{figure*}
    \centering
    \includegraphics[width=\linewidth]{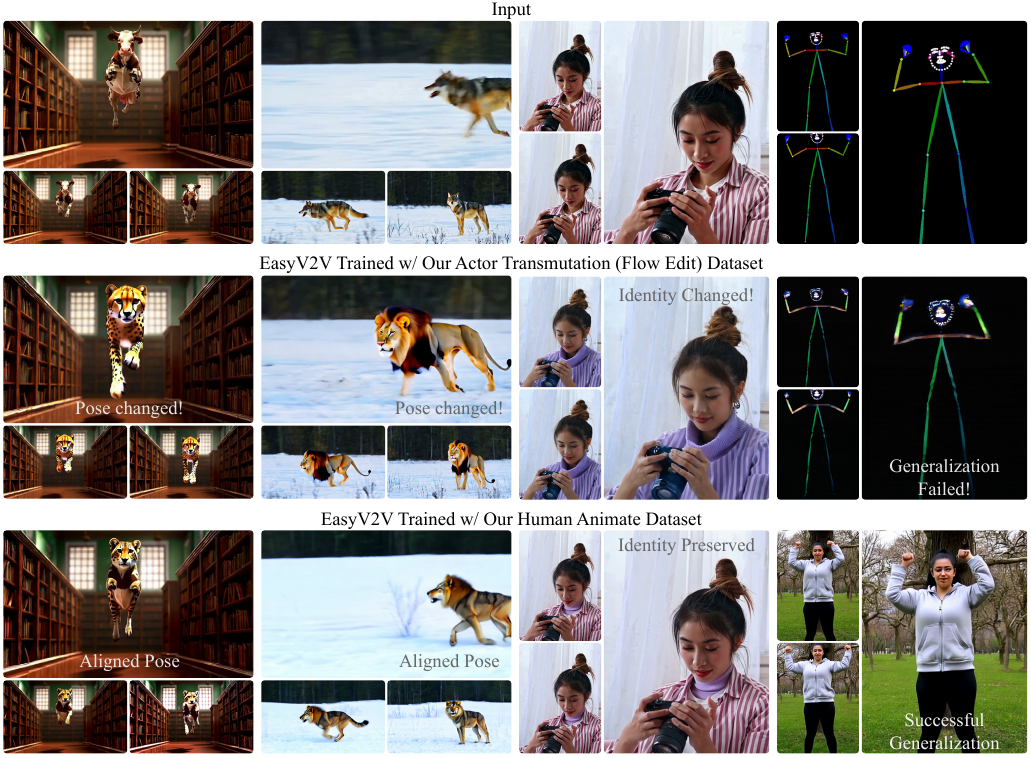}
    \caption{Comparison between models trained on the Flow Edit dataset and the human animate dataset.
}
    \label{fig:flowedit_vs_humananimate}
\end{figure*}

We curate the Human Animate dataset, which contains human-centric video edits, and the Flow Edit dataset, which focuses on actor transmutation edits.
We provide comparisons in Figure~\ref{fig:flowedit_vs_humananimate} between a model trained exclusively on the Human Animate dataset and one trained on the Flow Edit dataset.
The model trained on the Human Animate dataset often achieves superior visual details, preserves poses more effectively, and generalizes better to unseen human-specific pose-to-video tasks.
Facial expressions are also better preserved when training on the Human Animate dataset compared to the Flow Edit (actor transmutation) dataset.

\subsection{Exhibition Gallery and Visualization}

We present additional video editing examples on our visualization website included in the supplementary material.
We strongly encourage readers to view the attached HTML file for comprehensive and diverse video results.

\mysection{Capability for Human Action Editing}
Leveraging a densely captioned video dataset during training, our model demonstrates a strong capability to follow text instructions for modifying human actions.
Compared to concurrent works, which often struggle to alter human actions effectively, EasyV2V exhibits a \textbf{unique} proficiency in such edits, highlighting the effectiveness of our curated human action data.

\mysection{Natural Edit Transitions}
Although we employ a simple blending transition strategy during training, EasyV2V exhibits an emergent ability to produce natural transitions.
We evaluate this using frame-wise temporal masks where the edit is restricted to the second half of the video.
We observe smooth and realistic transitions at the timestamp where the source-to-target edit initiates.

\mysection{High-Resolution Results}
We train our model at a higher resolution of $81 \times 1280 \times 704$ to further validate our method and data pipeline.
The total training data is subsampled to approximately 6 million samples due to the computational cost of high-resolution training.
Note that $81 \times 1280 \times 704$ is the maximum supported resolution of Wan-2.2-TI2V-5B~\cite{wan2025wanopenadvancedlargescale}.
We observe rapid convergence within a few training steps, confirming that our architecture design with low-rank tuning efficiently adapts a T2V model to V2V tasks.
To the best of our knowledge, EasyV2V is the first instruction-based video editing model capable of editing $\sim$720P videos with a duration of 81 frames.

\section{VLM prompts for evaluation}
VLM prompt we used for spatial mask editing evaluation:
\begin{vlmprompt}
'You are a meticulous video editing quality evaluator. Your task is to provide a detailed assessment of a video edit by comparing the original image with the edited image based on a given text prompt.\n\
Editing Prompt:\n{editing_prompt}\n\
Instructions:\n\
Analyze the provided image (the edited video frame) and evaluate how well the "Editing Prompt" has been executed. You will evaluate the edit across three distinct criteria. For each criterion, provide a score from 0 (worst) to 3 (best) and a brief justification. Finally, provide the total score.\n\
Your evaluation should focus on three key aspects:\n\
1. Prompt Following (Score: 0-3) \n\
Question: Does the edit accurately and completely fulfill the instructions in the "Editing Prompt"? \n\
Scoring Guide:\n\
- 3: The prompt is perfectly and completely followed.\n\
- 2: The prompt is mostly followed but with minor inaccuracies or omissions.\n\
- 1: The prompt is poorly followed or only partially executed.\n\
- 0: The prompt is completely ignored or the opposite was done. \n\
2. Edit Quality (Score: 0-3) \n\
Question: How is the visual quality of the edited area itself? Is it realistic, seamless, and free of artifacts (e.g., blurriness, distortion, unnatural textures)?\n\
Scoring Guide:\n\
- 3: The edit is of high visual quality, seamless, and artifact-free.\n\
- 2: The edit is good but has minor, noticeable artifacts.\n\
- 1: The edit is of low quality with significant, distracting artifacts.\n\
- 0: The edited area is extremely poor, garbled, or has completely failed.\n\
3. Background Consistency (Score: 0-3) \n\
Question: Have the areas that should not have been edited remained unchanged between the "Before" and "After" images? \n\
Scoring Guide:\n\
- 3: The areas that should not have been edited are perfectly preserved and stable. \n\
- 2: There are minor, subtle, but noticeable changes or flickers in the areas that should not have been edited.\n\
- 1: There are significant and distracting changes in the areas that should not have been edited. \n\
- 0: The areas that should not have been edited is completely or catastrophically altered. \n\
\end{vlmprompt}

VLM prompt we used for temporal mask editing evaluation:
\begin{vlmprompt}
You are a meticulous video editing quality evaluator. Your task is to provide a detailed assessment of a temporal video edit by comparing the original image with the edited image based on a given text prompt.

Editing Prompt:
{editing_prompt}

IMPORTANT TEMPORAL CONTEXT:
This frame is from the FIRST HALF of the video, where the edit SHOULD NOT have occurred yet. The frame should remain unchanged from the original.

....
Your evaluation should focus on three key aspects:

1. Temporal Consistency (Score: 0-3)
Question: Does the frame correctly show NO editing in the first half of the video?
Scoring Guide:
- 3: The frame is perfectly unchanged, showing no signs of the edit that should only appear in the second half.
- 2: The frame is mostly unchanged but shows very minor, subtle hints of the edit.
- 1: The frame shows partial editing when it should be unchanged.
- 0: The frame is fully edited when it should be completely unchanged.

#############
IMPORTANT TEMPORAL CONTEXT:
This frame is from the TRANSITION PERIOD (around the middle of the video), where the edit IS HAPPENING. The frame should show the edit in progress or just completed.

Instructions:
Analyze the provided images and evaluate how well the temporal editing is progressing. You will evaluate the edit across three distinct criteria. For each criterion, provide a score from 0 (worst) to 3 (best) and a brief justification. Finally, provide the total score.

Your evaluation should focus on three key aspects:

1. Edit Progress (Score: 0-3)
Question: Does the frame show appropriate edit progression during this transition period?
Scoring Guide:
- 3: The frame shows natural, smooth editing transition that aligns with the temporal position.
- 2: The frame shows editing but the transition is slightly abrupt or unnatural.
- 1: The frame shows poor editing progression or timing.
- 0: The frame shows no editing or completely wrong timing.

....
IMPORTANT TEMPORAL CONTEXT:
This frame is from the SECOND HALF of the video, where the edit SHOULD HAVE been completed. The frame should show the fully edited result.

Instructions:
Analyze the provided images and evaluate how well the "Editing Prompt" has been executed. You will evaluate the edit across three distinct criteria. For each criterion, provide a score from 0 (worst) to 3 (best) and a brief justification. Finally, provide the total score.

Your evaluation should focus on three key aspects:

1. Prompt Following (Score: 0-3)
Question: Does the edit accurately and completely fulfill the instructions in the "Editing Prompt"?
Scoring Guide:
- 3: The prompt is perfectly and completely followed.
- 2: The prompt is mostly followed but with minor inaccuracies or omissions.
- 1: The prompt is poorly followed or only partially executed.
- 0: The prompt is completely ignored or the opposite was done.
#############

\end{vlmprompt}

{
    \small
    \bibliographystyle{ieeenat_fullname}
    \bibliography{main}

\begin{thebibliography}{73}
\providecommand{\natexlab}[1]{#1}
\providecommand{\url}[1]{\texttt{#1}}
\expandafter\ifx\csname urlstyle\endcsname\relax
  \providecommand{\doi}[1]{doi: #1}\else
  \providecommand{\doi}{doi: \begingroup \urlstyle{rm}\Url}\fi

\bibitem[Vid(2025)]{VideoX-Fun2025}
{VideoX-Fun}: A more flexible framework that can generate videos at any resolution and creates videos from images.
\newblock \url{https://github.com/aigc-apps/VideoX-Fun}, 2025.

\bibitem[Bai et~al.(2025{\natexlab{a}})Bai, Wang, Ouyang, Yu, Wang, Wang, Cheng, Ma, Zeng, Liu, Xu, Shen, and Chen]{bai2025ditto}
Qingyan Bai, Qiuyu Wang, Hao Ouyang, Yue Yu, Hanlin Wang, Wen Wang, Ka~Leong Cheng, Shuailei Ma, Yanhong Zeng, Zichen Liu, Yinghao Xu, Yujun Shen, and Qifeng Chen.
\newblock Scaling instruction-based video editing with a high-quality synthetic dataset.
\newblock \emph{arXiv}, 2025{\natexlab{a}}.

\bibitem[Bai et~al.(2025{\natexlab{b}})Bai, Chen, Liu, Wang, Ge, Song, Dang, Wang, Wang, Tang, Zhong, Zhu, Yang, Li, Wan, Wang, Ding, Fu, Xu, Ye, Zhang, Xie, Cheng, Zhang, Yang, Xu, and Lin]{bai2025qwen25vltechnicalreport}
Shuai Bai, Keqin Chen, Xuejing Liu, Jialin Wang, Wenbin Ge, Sibo Song, Kai Dang, Peng Wang, Shijie Wang, Jun Tang, Humen Zhong, Yuanzhi Zhu, Mingkun Yang, Zhaohai Li, Jianqiang Wan, Pengfei Wang, Wei Ding, Zheren Fu, Yiheng Xu, Jiabo Ye, Xi Zhang, Tianbao Xie, Zesen Cheng, Hang Zhang, Zhibo Yang, Haiyang Xu, and Junyang Lin.
\newblock {Qwen2.5-VL} technical report, 2025{\natexlab{b}}.

\bibitem[Brooks et~al.(2023)Brooks, Holynski, and Efros]{brooks2023instructpix2pix}
Tim Brooks, Aleksander Holynski, and Alexei~A Efros.
\newblock {InstructPix2Pix}: Learning to follow image editing instructions.
\newblock \emph{CVPR}, 2023.

\bibitem[Brooks et~al.(2024)Brooks, Peebles, Holmes, DePue, Guo, Jing, Schnurr, Taylor, Luhman, Luhman, Ng, Wang, and Ramesh]{videoworldsimulators2024}
Tim Brooks, Bill Peebles, Connor Holmes, Will DePue, Yufei Guo, Li Jing, David Schnurr, Joe Taylor, Troy Luhman, Eric Luhman, Clarence Ng, Ricky Wang, and Aditya Ramesh.
\newblock Video generation models as world simulators.
\newblock 2024.

\bibitem[Cai et~al.(2025)Cai, Chen, Chen, Li, Long, Pan, Qiu, Zhang, Gao, Xu, et~al.]{hidreami1technicalreport}
Qi Cai, Jingwen Chen, Yang Chen, Yehao Li, Fuchen Long, Yingwei Pan, Zhaofan Qiu, Yiheng Zhang, Fengbin Gao, Peihan Xu, et~al.
\newblock {HiDream-I1}: A high-efficient image generative foundation model with sparse diffusion transformer.
\newblock \emph{arXiv}, 2025.

\bibitem[Cao et~al.(2023)Cao, Wang, Qi, Shan, Qie, and Zheng]{cao2023masactrl}
Mingdeng Cao, Xintao Wang, Zhongang Qi, Ying Shan, Xiaohu Qie, and Yinqiang Zheng.
\newblock {MasaCtrl}: Tuning-free mutual self-attention control for consistent image synthesis and editing.
\newblock In \emph{ICCV}, 2023.

\bibitem[Cheng et~al.(2025)Cheng, Gao, Hu, Hu, Huang, Ji, Li, Meng, Qi, Qiao, Shen, Song, Sun, Tian, Wang, Wang, Wang, Wang, Xiao, Xu, Zhang, Zhang, Zhang, Zhang, Zhou, and Zhuo]{wananimate2025}
Gang Cheng, Xin Gao, Li Hu, Siqi Hu, Mingyang Huang, Chaonan Ji, Ju Li, Dechao Meng, Jinwei Qi, Penchong Qiao, Zhen Shen, Yafei Song, Ke Sun, Linrui Tian, Feng Wang, Guangyuan Wang, Qi Wang, Zhongjian Wang, Jiayu Xiao, Sheng Xu, Bang Zhang, Peng Zhang, Xindi Zhang, Zhe Zhang, Jingren Zhou, and Lian Zhuo.
\newblock {Wan-Animate}: Unified character animation and replacement with holistic replication, 2025.

\bibitem[Cheng et~al.(2024)Cheng, Xiao, and He]{cheng2023consistentvideotovideotransferusing}
Jiaxin Cheng, Tianjun Xiao, and Tong He.
\newblock Consistent video-to-video transfer using synthetic dataset.
\newblock \emph{ICLR}, 2024.

\bibitem[Couairon et~al.(2024)Couairon, Rambour, HAUGEARD, and THOME]{couairon2024videdit}
Paul Couairon, Cl{\'e}ment Rambour, Jean-Emmanuel HAUGEARD, and Nicolas THOME.
\newblock {VidEdit}: Zero-shot and spatially aware text-driven video editing.
\newblock \emph{TMLR}, 2024.

\bibitem[Deng et~al.(2025)Deng, Zhu, Li, Gou, and Li]{deng2025bagel}
Chaorui Deng, Deyao Zhu, Kunchang Li, Chenhui Gou, and Feng Li.
\newblock Emerging properties in unified multimodal pretraining: The {BAGEL} model.
\newblock \emph{arXiv}, 2025.

\bibitem[Gao et~al.(2025)Gao, Gong, Guo, Hou, Lai, Li, Li, Lian, Liao, Liu, Liu, Shi, Sun, Tian, Tian, Wang, Wang, Wang, Wang, Wang, Wu, Wu, Xia, Xiao, Zhai, Zhang, Zhang, Zhang, Zhao, Yang, and Huang]{gao2025seedream30technicalreport}
Yu Gao, Lixue Gong, Qiushan Guo, Xiaoxia Hou, Zhichao Lai, Fanshi Li, Liang Li, Xiaochen Lian, Chao Liao, Liyang Liu, Wei Liu, Yichun Shi, Shiqi Sun, Yu Tian, Zhi Tian, Peng Wang, Rui Wang, Xuanda Wang, Xun Wang, Ye Wang, Guofeng Wu, Jie Wu, Xin Xia, Xuefeng Xiao, Zhonghua Zhai, Xinyu Zhang, Qi Zhang, Yuwei Zhang, Shijia Zhao, Jianchao Yang, and Weilin Huang.
\newblock {Seedream 3.0} technical report, 2025.

\bibitem[Ge et~al.(2024)Ge, Zhao, Li, Ge, and Shan]{ge2024seed}
Yuying Ge, Sijie Zhao, Chen Li, Yixiao Ge, and Ying Shan.
\newblock {SEED-Data-Edit} technical report: A hybrid dataset for instructional image editing.
\newblock \emph{arXiv}, 2024.

\bibitem[Hertz et~al.(2022)Hertz, Mokady, Tenenbaum, Aberman, Pritch, and Cohen-Or]{hertz2022prompt}
Amir Hertz, Ron Mokady, Jay Tenenbaum, Kfir Aberman, Yael Pritch, and Daniel Cohen-Or.
\newblock Prompt-to-prompt image editing with cross attention control.
\newblock \emph{arXiv}, 2022.

\bibitem[Ho and Salimans(2022)]{CFG}
Jonathan Ho and Tim Salimans.
\newblock Classifier-free diffusion guidance.
\newblock \emph{arXiv preprint arXiv:2207.12598}, 2022.

\bibitem[Hu et~al.(2022)Hu, Shen, Wallis, Allen-Zhu, Li, Wang, Wang, and Chen]{hu2022lora}
Edward~J Hu, Yelong Shen, Phillip Wallis, Zeyuan Allen-Zhu, Yuanzhi Li, Shean Wang, Lu Wang, and Weizhu Chen.
\newblock Lo{RA}: Low-rank adaptation of large language models.
\newblock In \emph{ICLR}, 2022.

\bibitem[Hu et~al.(2025)Hu, Zhong, Wang, Jiang, Tian, Yang, Wan, and Zhang]{hu2025vivid10mdatasetbaselineversatile}
Jiahao Hu, Tianxiong Zhong, Xuebo Wang, Boyuan Jiang, Xingye Tian, Fei Yang, Pengfei Wan, and Di Zhang.
\newblock {VIVID-10M}: A dataset and baseline for versatile and interactive video local editing, 2025.

\bibitem[Hui et~al.(2024)Hui, Yang, Zhao, Shi, Wang, Wang, Zhou, and Xie]{hui2024hqedit}
Mude Hui, Siwei Yang, Bingchen Zhao, Yichun Shi, Heng Wang, Peng Wang, Yuyin Zhou, and Cihang Xie.
\newblock {HQ-Edit}: A high-quality dataset for instruction-based image editing.
\newblock \emph{arXiv}, 2024.

\bibitem[Imagen-Team-Google(2024)]{imagenteamgoogle2024imagen3}
Imagen-Team-Google.
\newblock {Imagen 3}, 2024.

\bibitem[Jiang et~al.(2025)Jiang, Han, Mao, Zhang, Pan, and Liu]{vace}
Zeyinzi Jiang, Zhen Han, Chaojie Mao, Jingfeng Zhang, Yulin Pan, and Yu Liu.
\newblock {VACE}: All-in-one video creation and editing.
\newblock In \emph{ICCV}, 2025.

\bibitem[Ju et~al.(2024)Ju, Zeng, Bian, Liu, and Xu]{ju2023direct}
Xuan Ju, Ailing Zeng, Yuxuan Bian, Shaoteng Liu, and Qiang Xu.
\newblock {PnP} inversion: Boosting diffusion-based editing with 3 lines of code.
\newblock \emph{ICLR}, 2024.

\bibitem[Ju et~al.(2025{\natexlab{a}})Ju, Wang, Zhou, Zhang, Liu, Zhao, Zhang, Li, Cai, Liu, Pakhomov, Lin, Kim, and Xu]{editverse2025}
Xuan Ju, Tianyu Wang, Yuqian Zhou, He Zhang, Qing Liu, Nanxuan Zhao, Zhifei Zhang, Yijun Li, Yuanhao Cai, Shaoteng Liu, Daniil Pakhomov, Zhe Lin, Soo~Ye Kim, and Qiang Xu.
\newblock {EditVerse}: Unifying image and video editing and generation with {In-Context} learning, 2025{\natexlab{a}}.

\bibitem[Ju et~al.(2025{\natexlab{b}})Ju, Ye, Liu, Wang, Wang, Wan, Zhang, Gai, and Xu]{ju2025fullditmultitaskvideogenerative}
Xuan Ju, Weicai Ye, Quande Liu, Qiulin Wang, Xintao Wang, Pengfei Wan, Di Zhang, Kun Gai, and Qiang Xu.
\newblock {FullDiT}: Multi-task video generative foundation model with full attention, 2025{\natexlab{b}}.

\bibitem[Kirstain et~al.(2023)Kirstain, Polyak, Singer, Matiana, Penna, and Levy]{pickscore}
Yuval Kirstain, Adam Polyak, Uriel Singer, Shahbuland Matiana, Joe Penna, and Omer Levy.
\newblock Pick-a-pic: An open dataset of user preferences for text-to-image generation.
\newblock \emph{Advances in neural information processing systems}, 36:\penalty0 36652--36663, 2023.

\bibitem[Kong et~al.(2024)Kong, Tian, Zhang, Min, Dai, Zhou, Xiong, Li, Wu, Zhang, et~al.]{kong2024hunyuanvideo}
Weijie Kong, Qi Tian, Zijian Zhang, Rox Min, Zuozhuo Dai, Jin Zhou, Jiangfeng Xiong, Xin Li, Bo Wu, Jianwei Zhang, et~al.
\newblock {Hunyuanvideo}: A systematic framework for large video generative models.
\newblock \emph{arXiv}, 2024.

\bibitem[Ku et~al.(2024)Ku, Wei, Ren, Yang, and Chen]{ku2024anyv2v}
Max Ku, Cong Wei, Weiming Ren, Harry Yang, and Wenhu Chen.
\newblock {AnyV2V}: A tuning-free framework for any video-to-video editing tasks.
\newblock \emph{TMLR}, 2024.

\bibitem[Kulikov et~al.(2025)Kulikov, Kleiner, Huberman-Spiegelglas, and Michaeli]{kulikov2024flowedit}
Vladimir Kulikov, Matan Kleiner, Inbar Huberman-Spiegelglas, and Tomer Michaeli.
\newblock {FlowEdit}: Inversion-free text-based editing using pre-trained flow models.
\newblock \emph{ICCV}, 2025.

\bibitem[Labs(2024)]{flux2024}
Black~Forest Labs.
\newblock {FLUX}.
\newblock \url{https://github.com/black-forest-labs/flux}, 2024.

\bibitem[Labs et~al.(2025)Labs, Batifol, Blattmann, Boesel, Consul, Diagne, Dockhorn, English, English, Esser, Kulal, Lacey, Levi, Li, Lorenz, Müller, Podell, Rombach, Saini, Sauer, and Smith]{labs2025flux1kontextflowmatching}
Black~Forest Labs, Stephen Batifol, Andreas Blattmann, Frederic Boesel, Saksham Consul, Cyril Diagne, Tim Dockhorn, Jack English, Zion English, Patrick Esser, Sumith Kulal, Kyle Lacey, Yam Levi, Cheng Li, Dominik Lorenz, Jonas Müller, Dustin Podell, Robin Rombach, Harry Saini, Axel Sauer, and Luke Smith.
\newblock {FLUX.1 Kontext}: Flow matching for in-context image generation and editing in latent space, 2025.

\bibitem[Liu et~al.(2024)Liu, Zhang, Li, Lin, and Jia]{liu2024videop2p}
Shaoteng Liu, Yuechen Zhang, Wenbo Li, Zhe Lin, and Jiaya Jia.
\newblock {Video-P2P}: Video editing with cross-attention control.
\newblock \emph{CVPR}, 2024.

\bibitem[Liu et~al.(2025)Liu, Wang, Wang, Liu, Zhang, Lee, Li, Yu, Lin, Kim, and Jia]{liu2025genprop}
Shaoteng Liu, Tianyu Wang, Jui-Hsien Wang, Qing Liu, Zhifei Zhang, Joon-Young Lee, Yijun Li, Bei Yu, Zhe Lin, Soo~Ye Kim, and Jiaya Jia.
\newblock Generative video propagation.
\newblock \emph{CVPR}, 2025.

\bibitem[Loshchilov and Hutter(2017)]{adamw}
Ilya Loshchilov and Frank Hutter.
\newblock Decoupled weight decay regularization.
\newblock \emph{arXiv preprint arXiv:1711.05101}, 2017.

\bibitem[Qi et~al.(2023)Qi, Cun, Zhang, Lei, Wang, Shan, and Chen]{qi2023fatezero}
Chenyang Qi, Xiaodong Cun, Yong Zhang, Chenyang Lei, Xintao Wang, Ying Shan, and Qifeng Chen.
\newblock {FateZero}: Fusing attentions for zero-shot text-based video editing.
\newblock \emph{ICCV}, 2023.

\bibitem[Qin et~al.(2024)Qin, Li, Tang, Chua, and Zhuang]{InstructVid2Vid}
Bosheng Qin, Juncheng Li, Siliang Tang, Tat-Seng Chua, and Yueting Zhuang.
\newblock {InstructVid2Vid}: Controllable video editing with natural language instructions.
\newblock In \emph{ICME}, 2024.

\bibitem[Qu et~al.(2025)Qu, Zhang, Liu, Wang, Jiang, Gao, Ye, Du, Yuan, and Wu]{tokenflow}
Liao Qu, Huichao Zhang, Yiheng Liu, Xu Wang, Yi Jiang, Yiming Gao, Hu Ye, Daniel~K. Du, Zehuan Yuan, and Xinglong Wu.
\newblock Tokenflow: Unified image tokenizer for multimodal understanding and generation.
\newblock In \emph{CVPR}, 2025.

\bibitem[{Qwen Team}(2025)]{qwenimage2025}
{Qwen Team}.
\newblock {Qwen-Image}: A unified foundation model for image generation and editing.
\newblock \emph{arXiv}, 2025.

\bibitem[Radford et~al.(2021)Radford, Kim, Hallacy, Ramesh, Goh, Agarwal, Sastry, Askell, Mishkin, Clark, et~al.]{clip2021}
Alec Radford, Jong~Wook Kim, Chris Hallacy, Aditya Ramesh, Gabriel Goh, Sandhini Agarwal, Girish Sastry, Amanda Askell, Pamela Mishkin, Jack Clark, et~al.
\newblock Learning transferable visual models from natural language supervision.
\newblock In \emph{International conference on machine learning}, pages 8748--8763. PmLR, 2021.

\bibitem[Ravi et~al.(2024)Ravi, Gabeur, Hu, Hu, Ryali, Ma, Khedr, R{\"a}dle, Rolland, Gustafson, Mintun, Pan, Alwala, Carion, Wu, Girshick, Doll{\'a}r, and Feichtenhofer]{ravi2024sam2}
Nikhila Ravi, Valentin Gabeur, Yuan-Ting Hu, Ronghang Hu, Chaitanya Ryali, Tengyu Ma, Haitham Khedr, Roman R{\"a}dle, Chloe Rolland, Laura Gustafson, Eric Mintun, Junting Pan, Kalyan~Vasudev Alwala, Nicolas Carion, Chao-Yuan Wu, Ross Girshick, Piotr Doll{\'a}r, and Christoph Feichtenhofer.
\newblock {SAM 2}: Segment anything in images and videos.
\newblock \emph{arXiv}, 2024.

\bibitem[Rombach et~al.(2022)Rombach, Blattmann, Lorenz, Esser, and Ommer]{Rombach_2022_CVPR}
Robin Rombach, Andreas Blattmann, Dominik Lorenz, Patrick Esser, and Bj\"orn Ommer.
\newblock High-resolution image synthesis with latent diffusion models.
\newblock In \emph{CVPR}, 2022.

\bibitem[Runway()]{runway2025aleph}
Runway.
\newblock Introducing runway aleph.
\newblock \url{https://runwayml.com/research/introducing-runway-aleph}.

\bibitem[Seedream et~al.(2025)Seedream, :, Chen, Gao, Gong, Guo, Guo, Guo, Hou, Huang, Huang, Jian, Kuang, Lai, Li, Li, Lian, Liao, Liu, Liu, Lu, Luo, Ou, Shi, Shi, Sun, Tian, Tian, Wang, Wang, Wang, Wang, Wu, Wu, Wu, Wu, Xia, Xiao, Xu, Yan, Yang, Yang, Zhai, Zhang, Zhang, Zhang, Zhang, Zhang, Zhao, Zhao, and Zhu]{seedream2025seedream40nextgenerationmultimodal}
Team Seedream, :, Yunpeng Chen, Yu Gao, Lixue Gong, Meng Guo, Qiushan Guo, Zhiyao Guo, Xiaoxia Hou, Weilin Huang, Yixuan Huang, Xiaowen Jian, Huafeng Kuang, Zhichao Lai, Fanshi Li, Liang Li, Xiaochen Lian, Chao Liao, Liyang Liu, Wei Liu, Yanzuo Lu, Zhengxiong Luo, Tongtong Ou, Guang Shi, Yichun Shi, Shiqi Sun, Yu Tian, Zhi Tian, Peng Wang, Rui Wang, Xun Wang, Ye Wang, Guofeng Wu, Jie Wu, Wenxu Wu, Yonghui Wu, Xin Xia, Xuefeng Xiao, Shuang Xu, Xin Yan, Ceyuan Yang, Jianchao Yang, Zhonghua Zhai, Chenlin Zhang, Heng Zhang, Qi Zhang, Xinyu Zhang, Yuwei Zhang, Shijia Zhao, Wenliang Zhao, and Wenjia Zhu.
\newblock {Seedream 4.0}: Toward next-generation multimodal image generation, 2025.

\bibitem[Sheynin et~al.(2023)Sheynin, Polyak, Singer, Kirstain, Zohar, Ashual, Parikh, and Taigman]{sheynin2024emuedit}
Shelly Sheynin, Adam Polyak, Uriel Singer, Yuval Kirstain, Amit Zohar, Oron Ashual, Devi Parikh, and Yaniv Taigman.
\newblock Emu edit: Precise image editing via recognition and generation tasks.
\newblock In \emph{CVPR}, 2023.

\bibitem[Singer et~al.(2024)Singer, Zohar, Kirstain, Sheynin, Polyak, Parikh, and Taigman]{singer2024eve}
Uriel Singer, Amit Zohar, Yuval Kirstain, Shelly Sheynin, Adam Polyak, Devi Parikh, and Yaniv Taigman.
\newblock {EVE}: Video editing via factorized diffusion distillation.
\newblock \emph{ECCV}, 2024.

\bibitem[Team(2025)]{lucyedit2024}
Decart Team.
\newblock Lucy edit: Open-weight text-guided video editing.
\newblock \url{https://d2drjpuinn46lb.cloudfront.net/Lucy_Edit__High_Fidelity_Text_Guided_Video_Editing.pdf}, 2025.

\bibitem[Team(2024{\natexlab{a}})]{genmo2024mochi}
Genmo Team.
\newblock Mochi 1.
\newblock \url{https://github.com/genmoai/models}, 2024{\natexlab{a}}.

\bibitem[Team et~al.(2025)Team, Kamath, Ferret, Pathak, Vieillard, Merhej, Perrin, Matejovicova, Ram{\'e}, Rivi{\`e}re, et~al.]{gemma3}
Gemma Team, Aishwarya Kamath, Johan Ferret, Shreya Pathak, Nino Vieillard, Ramona Merhej, Sarah Perrin, Tatiana Matejovicova, Alexandre Ram{\'e}, Morgane Rivi{\`e}re, et~al.
\newblock Gemma 3 technical report.
\newblock \emph{arXiv preprint arXiv:2503.19786}, 2025.

\bibitem[Team(2024{\natexlab{b}})]{openai2024gpt4ocard}
OpenAI Team.
\newblock {GPT-4o} system card, 2024{\natexlab{b}}.

\bibitem[Wan et~al.(2025)Wan, Wang, Ai, Wen, Mao, Xie, Chen, Yu, Zhao, Yang, Zeng, Wang, Zhang, Zhou, Wang, Chen, Zhu, Zhao, Yan, Huang, Feng, Zhang, Li, Wu, Chu, Feng, Zhang, Sun, Fang, Wang, Gui, Weng, Shen, Lin, Wang, Wang, Zhou, Wang, Shen, Yu, Shi, Huang, Xu, Kou, Lv, Li, Liu, Wang, Zhang, Huang, Li, Wu, Liu, Pan, Zheng, Hong, Shi, Feng, Jiang, Han, Wu, and Liu]{wan2025wanopenadvancedlargescale}
Team Wan, Ang Wang, Baole Ai, Bin Wen, Chaojie Mao, Chen-Wei Xie, Di Chen, Feiwu Yu, Haiming Zhao, Jianxiao Yang, Jianyuan Zeng, Jiayu Wang, Jingfeng Zhang, Jingren Zhou, Jinkai Wang, Jixuan Chen, Kai Zhu, Kang Zhao, Keyu Yan, Lianghua Huang, Mengyang Feng, Ningyi Zhang, Pandeng Li, Pingyu Wu, Ruihang Chu, Ruili Feng, Shiwei Zhang, Siyang Sun, Tao Fang, Tianxing Wang, Tianyi Gui, Tingyu Weng, Tong Shen, Wei Lin, Wei Wang, Wei Wang, Wenmeng Zhou, Wente Wang, Wenting Shen, Wenyuan Yu, Xianzhong Shi, Xiaoming Huang, Xin Xu, Yan Kou, Yangyu Lv, Yifei Li, Yijing Liu, Yiming Wang, Yingya Zhang, Yitong Huang, Yong Li, You Wu, Yu Liu, Yulin Pan, Yun Zheng, Yuntao Hong, Yupeng Shi, Yutong Feng, Zeyinzi Jiang, Zhen Han, Zhi-Fan Wu, and Ziyu Liu.
\newblock {Wan}: Open and advanced large-scale video generative models, 2025.

\bibitem[Wang et~al.(2023{\natexlab{a}})Wang, Jiang, Xie, Liu, Chen, Cao, Wang, and Shen]{wang2024vid2vidzero}
Wen Wang, Yan Jiang, Kangyang Xie, Zide Liu, Hao Chen, Yue Cao, Xinlong Wang, and Chunhua Shen.
\newblock Zero-shot video editing using off-the-shelf image diffusion models.
\newblock \emph{arXiv}, 2023{\natexlab{a}}.

\bibitem[Wang et~al.(2023{\natexlab{b}})Wang, He, Li, Li, Yu, Ma, Li, Chen, Chen, Wang, et~al.]{viclip}
Yi Wang, Yinan He, Yizhuo Li, Kunchang Li, Jiashuo Yu, Xin Ma, Xinhao Li, Guo Chen, Xinyuan Chen, Yaohui Wang, et~al.
\newblock Internvid: A large-scale video-text dataset for multimodal understanding and generation.
\newblock \emph{arXiv preprint arXiv:2307.06942}, 2023{\natexlab{b}}.

\bibitem[Wang et~al.(2025)Wang, Yang, Zhao, Zhang, Liu, Zhou, and Xie]{gpt_image_edit}
Yuhan Wang, Siwei Yang, Bingchen Zhao, Letian Zhang, Qing Liu, Yuyin Zhou, and Cihang Xie.
\newblock {GPT-IMAGE-EDIT-1.5M}: A million-scale, {GPT}-generated image dataset, 2025.

\bibitem[Wei et~al.(2025)Wei, Xiong, Ren, Du, Zhang, and Chen]{wei2025omniedit}
Cong Wei, Zheyang Xiong, Weiming Ren, Xinrun Du, Ge Zhang, and Wenhu Chen.
\newblock {OmniEdit}: Building image editing generalist models through specialist supervision.
\newblock In \emph{ICLR}, 2025.

\bibitem[Wu et~al.(2025{\natexlab{a}})Wu, Chen, Li, Wang, Xie, and Zhang]{wu2025insvie}
Yuhui Wu, Liyi Chen, Ruibin Li, Shihao Wang, Chenxi Xie, and Lei Zhang.
\newblock {InsViE-1M}: Effective instruction-based video editing with elaborate dataset construction.
\newblock \emph{ICCV}, 2025{\natexlab{a}}.

\bibitem[Wu et~al.(2025{\natexlab{b}})Wu, Siarohin, Menapace, Skorokhodov, Fang, Chordia, Gilitschenski, and Tulyakov]{mint}
Ziyi Wu, Aliaksandr Siarohin, Willi Menapace, Ivan Skorokhodov, Yuwei Fang, Varnith Chordia, Igor Gilitschenski, and Sergey Tulyakov.
\newblock Mind the time: Temporally-controlled multi-event video generation.
\newblock In \emph{Proceedings of the Computer Vision and Pattern Recognition Conference}, pages 23989--24000, 2025{\natexlab{b}}.

\bibitem[Xiao et~al.(2025)Xiao, Wang, Zhou, Yuan, Xing, Yan, Li, Wang, Huang, and Liu]{xiao2025omnigen}
Shitao Xiao, Yueze Wang, Junjie Zhou, Huaying Yuan, Xingrun Xing, Ruiran Yan, Chaofan Li, Shuting Wang, Tiejun Huang, and Zheng Liu.
\newblock {OmniGen}: Unified image generation.
\newblock In \emph{CVPR}, 2025.

\bibitem[Yang et~al.(2025{\natexlab{a}})Yang, Li, Yang, Zhang, Hui, Zheng, Yu, Gao, Huang, Lv, et~al.]{qwen3}
An Yang, Anfeng Li, Baosong Yang, Beichen Zhang, Binyuan Hui, Bo Zheng, Bowen Yu, Chang Gao, Chengen Huang, Chenxu Lv, et~al.
\newblock {Qwen3} technical report.
\newblock \emph{arXiv}, 2025{\natexlab{a}}.

\bibitem[Yang et~al.(2024)Yang, Zeng, Liu, Li, Xu, Zhang, and Yan]{yang2024editworld}
Ling Yang, Bohan Zeng, Jiaming Liu, Hong Li, Minghao Xu, Wentao Zhang, and Shuicheng Yan.
\newblock {EditWorld}: Simulating world dynamics for instruction-following image editing.
\newblock \emph{arXiv}, 2024.

\bibitem[Yang et~al.(2025{\natexlab{b}})Yang, Teng, Zheng, Ding, Huang, Xu, Yang, Hong, Zhang, Feng, et~al.]{yang2024cogvideox}
Zhuoyi Yang, Jiayan Teng, Wendi Zheng, Ming Ding, Shiyu Huang, Jiazheng Xu, Yuanming Yang, Wenyi Hong, Xiaohan Zhang, Guanyu Feng, et~al.
\newblock {CogVideoX}: Text-to-video diffusion models with an expert transformer.
\newblock \emph{ICLR}, 2025{\natexlab{b}}.

\bibitem[Yatim et~al.(2024)Yatim, Fridman, Bar-Tal, Kasten, and Dekel]{STDF}
Danah Yatim, Rafail Fridman, Omer Bar-Tal, Yoni Kasten, and Tali Dekel.
\newblock Space-time diffusion features for zero-shot text-driven motion transfer.
\newblock In \emph{CVPR}, 2024.

\bibitem[Ye et~al.(2025{\natexlab{a}})Ye, He, Li, Lin, Yuan, Yan, Hou, and Yuan]{ye2025imgeditbench}
Yang Ye, Xianyi He, Zongjian Li, Bin Lin, Shenghai Yuan, Zhiyuan Yan, Bohan Hou, and Li Yuan.
\newblock Imgedit: A unified image editing dataset and benchmark.
\newblock \emph{arXiv preprint arXiv:2505.20275}, 2025{\natexlab{a}}.

\bibitem[Ye et~al.(2025{\natexlab{b}})Ye, He, Liu, Wang, Wang, Wan, Zhang, Gai, Chen, and Luo]{ye2025unic}
Zixuan Ye, Xuanhua He, Quande Liu, Qiulin Wang, Xintao Wang, Pengfei Wan, Di Zhang, Kun Gai, Qifeng Chen, and Wenhan Luo.
\newblock {UNIC}: Unified in-context video editing, 2025{\natexlab{b}}.

\bibitem[Yoon et~al.(2025)Yoon, Yu, and Bansal]{yoon2025raccoon}
Jaehong Yoon, Shoubin Yu, and Mohit Bansal.
\newblock {RACCooN}: A versatile instructional video editing framework with auto-generated narratives.
\newblock \emph{EMNLP}, 2025.

\bibitem[Yu et~al.(2025{\natexlab{a}})Yu, Chow, Yue, Pan, Wu, Wan, Li, Tang, Zhang, and Zhuang]{yu2025anyedit}
Qifan Yu, Wei Chow, Zhongqi Yue, Kaihang Pan, Yang Wu, Xiaoyang Wan, Juncheng Li, Siliang Tang, Hanwang Zhang, and Yueting Zhuang.
\newblock {AnyEdit}: Mastering unified high-quality image editing for any idea.
\newblock In \emph{CVPR}, 2025{\natexlab{a}}.

\bibitem[Yu et~al.(2025{\natexlab{b}})Yu, Liu, Ma, Hong, Zhou, Tan, Chai, and Bansal]{yu2025veggie}
Shoubin Yu, Difan Liu, Ziqiao Ma, Yicong Hong, Yang Zhou, Hao Tan, Joyce Chai, and Mohit Bansal.
\newblock {VEGGIE}: Instructional editing and reasoning video concepts with grounded generation.
\newblock \emph{ICCV}, 2025{\natexlab{b}}.

\bibitem[Zhang et~al.(2023{\natexlab{a}})Zhang, Mo, Chen, Sun, and Su]{magicbrush2023}
Kai Zhang, Lingbo Mo, Wenhu Chen, Huan Sun, and Yu Su.
\newblock {MagicBrush}: A manually annotated dataset for instruction-guided image editing.
\newblock In \emph{NeurIPS}, 2023{\natexlab{a}}.

\bibitem[Zhang et~al.(2023{\natexlab{b}})Zhang, Rao, and Agrawala]{zhang2023adding}
Lvmin Zhang, Anyi Rao, and Maneesh Agrawala.
\newblock Adding conditional control to text-to-image diffusion models, 2023{\natexlab{b}}.

\bibitem[Zhang et~al.(2023{\natexlab{c}})Zhang, Huang, Ma, Li, Luo, Xie, Qin, Luo, Li, Liu, et~al.]{zhang2023recognize}
Youcai Zhang, Xinyu Huang, Jinyu Ma, Zhaoyang Li, Zhaochuan Luo, Yanchun Xie, Yuzhuo Qin, Tong Luo, Yaqian Li, Shilong Liu, et~al.
\newblock Recognize anything: A strong image tagging model.
\newblock \emph{arXiv}, 2023{\natexlab{c}}.

\bibitem[Zhang et~al.(2024)Zhang, Dai, Qin, and Wang]{zhang2024effived}
Zhenghao Zhang, Zuozhuo Dai, Long Qin, and Weizhi Wang.
\newblock {EffiVED}: Efficient video editing via text-instruction diffusion models, 2024.

\bibitem[Zhang et~al.(2025)Zhang, Xie, Lu, Yang, and Yang]{zhang2025icedit}
Zechuan Zhang, Ji Xie, Yu Lu, Zongxin Yang, and Yi Yang.
\newblock {In-Context Edit}: Enabling instructional image editing with in-context generation in large-scale diffusion transformers.
\newblock \emph{NeurIPS}, 2025.

\bibitem[Zhao et~al.(2024)Zhao, Ma, Chen, Si, Wu, An, Yu, Zhang, Li, and Chang]{zhao2024ultraedit}
Haozhe Zhao, Xiaojian Ma, Liang Chen, Shuzheng Si, Rujie Wu, Kaikai An, Peiyu Yu, Minjia Zhang, Qing Li, and Baobao Chang.
\newblock {UltraEdit}: Instruction-based fine-grained image editing at scale.
\newblock In \emph{NeurIPS}, 2024.

\bibitem[Zhou et~al.(2024)Zhou, Yu, Babu, Tirumala, Yasunaga, Shamis, Kahn, Ma, Zettlemoyer, and Levy]{zhou2024transfusionpredicttokendiffuse}
Chunting Zhou, Lili Yu, Arun Babu, Kushal Tirumala, Michihiro Yasunaga, Leonid Shamis, Jacob Kahn, Xuezhe Ma, Luke Zettlemoyer, and Omer Levy.
\newblock {Transfusion}: Predict the next token and diffuse images with one multi-modal model, 2024.

\bibitem[Zi et~al.(2025{\natexlab{a}})Zi, Peng, Qi, Wang, Zhao, Xiao, and Wong]{zi2025minimax}
Bojia Zi, Weixuan Peng, Xianbiao Qi, Jianan Wang, Shihao Zhao, Rong Xiao, and Kam-Fai Wong.
\newblock {MiniMax-Remover}: Taming bad noise helps video object removal.
\newblock \emph{arXiv}, 2025{\natexlab{a}}.

\bibitem[Zi et~al.(2025{\natexlab{b}})Zi, Ruan, Chen, Qi, Hao, Zhao, Huang, Liang, Xiao, and Wong]{zi2025senorita}
Bojia Zi, Penghui Ruan, Marco Chen, Xianbiao Qi, Shaozhe Hao, Shihao Zhao, Youze Huang, Bin Liang, Rong Xiao, and Kam-Fai Wong.
\newblock {Señorita-2M}: A high-quality instruction-based dataset for general video editing by video specialists.
\newblock In \emph{NeurIPS}, 2025{\natexlab{b}}.

\end{thebibliography}
}

\clearpage

\end{document}